\documentclass{article}
\usepackage{algpseudocode}
\usepackage{PRIMEarxiv}
\usepackage{algorithm}
\usepackage[utf8]{inputenc} 
\usepackage[T1]{fontenc}    
\usepackage{float}
\usepackage{hyperref}       
\usepackage{url}            
\usepackage{booktabs}       
\usepackage{amsfonts}       
\usepackage{amsmath}
\usepackage{amssymb}
\usepackage{nicefrac}       
\usepackage{microtype}      
\usepackage{fancyhdr}       
\usepackage{graphicx}       
\usepackage{tabularx}
\usepackage{array}
\usepackage{listings}
\usepackage{multicol}
\usepackage{placeins}
\usepackage{xcolor}
\usepackage[numbers,sort&compress]{natbib}
\graphicspath{{images/}}

\hypersetup{colorlinks=true, linkcolor=blue!60!black, citecolor=blue!60!black, urlcolor=blue!60!black}

\lstdefinestyle{advisory}{
  basicstyle=\ttfamily\fontsize{7}{8}\selectfont,
  columns=fullflexible,
  keepspaces=true,
  frame=single,
  rulecolor=\color{black!30},
  backgroundcolor=\color{black!3},
  xleftmargin=2pt, xrightmargin=2pt,
  aboveskip=4pt, belowskip=4pt,
  breaklines=true,
  breakindent=0pt,
  showstringspaces=false,
  morecomment=[l]{\#},
  commentstyle=\color{green!40!black}\bfseries,
  captionpos=b,
}
\lstnewenvironment{advisory}[1][]{\lstset{style=advisory,#1}}{}

\title{Agri-SCHOLAR: Simulation-Grounded In-Season Rolling Advisories for Agricultural Crops Using LLM-Bounded Optimization}

\author{
  Geetha Charan, Vedant Balasubramaniam,  Manojkumar Patil, V. Priyanka \\
  Indian Institute of Science \\
  Bengaluru, India \\
  \texttt{\{geethacharan, vedantb, pmanojkumar, priyankav\}@iisc.ac.in} \\
  \And
  Kodur Sai Vinay Sathvik \\
  Indian Institute of Science \\
  Bengaluru, India \\
  \texttt{sathvik2004@gmail.com} \\
  \And
  Rohit P Suresh \\
  University of Michigan \\
  Ann Arbor, USA \\
  \texttt{rohitps@umich.edu} \\
  \And
  Mohammad Sufyan Azam, Y. Narahari \\
  Indian Institute of Science \\
  Bengaluru, India \\
\texttt{\{mohammadazam, narahari\}@iisc.ac.in} \\
}

\begin{document}
\maketitle

\begin{abstract}
Existing agricultural advisories to farmers suffer from several limitations: (a) remain largely expert-driven, (b) do not adequately account for in-season weather, and (c) do not scale to individual farms. In this paper, our objective is to close these gaps by developing \textbf{Agri-SCHOLAR} (\textbf{S}imulation-Grounded \textbf{C}rop \textbf{H}arvest \textbf{O}ptimization with \textbf{L}LM-Guided In-Season \textbf{A}dvisory \textbf{R}ollouts). Agri-SCHOLAR is a rolling, in-season advisory framework that issues, at intervals of every few days or every week, a personalized advisory using the following information: soil and crop properties, observed weather to date, short-range forecast of weather, and a season extrapolation of weather based on historical analogs. The novelty and depth of Agri-SCHOLAR lies in each advisory being produced by a closed-loop system which couples multiple LLM agents with a powerful crop simulator (such as APSIM in our case). The simulator crucially serves as an evaluation oracle. In Agri-SCHOLAR, the LLM agents propose agronomically sound and feasible bounds on actionable parameters and an optimizer searches within those bounds against a multi-objective reward spanning yield, resource utilization, and economic feasibility. We conduct a detailed 11-year analysis of the maize crop in two selected locations of South Asia and show that Agri-SCHOLAR consistently outperforms the region's standard packages of practice and satisfactorily addresses the limitations of existing agricultural advisories.
\end{abstract}


\section{Introduction}

Agricultural advisories in much of the developing world are still produced by small teams of agronomists who issue broad guidance across large, dispersed regions. This approach does not scale, as even farms in the same advisory zone differ in soil, cultivar, sowing date, and management history, and they experience the season's weather differently. So a single recommendation is rarely optimal for an individual farm. Personalized, AI-driven advisories are a natural way to close this gap. For farmers to act on them, however, such advisories must be agronomically and physiologically feasible for the selected crop under the conditions of that particular farm.

Recent works \cite{balasubramaniam2026agri} address this requirement by using crop simulators like Decision Support System for Agrotechnology Transfer (DSSAT) and Agricultural Production Systems sIMulator (APSIM) to validate LLM-generated season-long advisories. This resolves the plausibility problem but introduces a new practical problem: evaluating a season-long plan in a simulator requires the weather for the entire season in advance. No forecasting system provides this at the required fidelity for a crop cycle, which can exceed several months, so such a pipeline cannot be deployed prospectively.

Operational advisory services avoid this problem by not planning the whole season at once. For example, India's Agro-Meteorological Field Units (AMFUs) issue advisories twice a week based on observed weather and short-range forecasts, and reissue them as the season unfolds; the agro-meteorological services of China and Vietnam follow similar practices. The underlying principle is that advice should be revised repeatedly as new weather information arrives. Existing operational advisories, however, remain expert-driven, are issued at the level of an administrative region rather than a farm, and are not checked against a mechanistic model of crop response.

Inspired by this practice, we introduce \textbf{Agri-SCHOLAR}, a rolling, in-season advisory framework that combines the simulator-based validation with the reissue-as-you-go principle of operational services. At each advisory step (every few days to a week), Agri-SCHOLAR evaluates candidate actions in APSIM using a weather trajectory assembled from three sources: (i) observed  weather to date, (ii) the available short-range forecast, and (iii) an extrapolation of the remainder of the season based on atmospherically analogous previous years. The advisory is then recomputed at the next step with updated observations of weather, so errors due to extrapolated weather in the previous step do not accumulate over the season.

A two-stage, closed-loop pipeline produces each advisory. In the first stage, LLM agents reason about the farm's soil, crop state, and current weather context to propose agronomically sound and feasible bounds on the actionable parameters (in this paper, the timing and amount of irrigation and fertilizer). In the second stage, an optimizer searches within these bounds to maximize a multi-objective reward that captures yield, nitrogen, and water use efficiency, and economic feasibility, with APSIM serving as the evaluation oracle for every candidate recommendation. This division of labor allows the LLM to perform semantic, context-dependent reasoning about which actions are appropriate for a given farm, while the simulator and optimizer quantitatively evaluate and compare those actions within the feasible decision space. The resulting advisories are therefore both farm-specific and constrained by the crop and environmental conditions under which they must remain feasible.

We evaluate Agri-SCHOLAR for maize in South Asia over 11 years, comparing it with the region's standard package of practices. Across seasons, Agri-SCHOLAR consistently outperforms this baseline on the reward components above, which indicates that rolling, simulator-grounded advisories can address the limitations of existing advisory systems without requiring an accurate season-long weather forecast.

\paragraph{\textbf{Contributions:}}
This paper makes the following contributions.
\begin{enumerate}
    \item  We introduce Agri-SCHOLAR, which is an eminently deployable, rolling, farm-specific crop advisory system that clearly improves upon classical season-long, generic, simulation-based advisories. Agri-SCHOLAR makes information constrained decision-making with weekly, farm-specific advisories produced by an LLM-bounded, simulation-validated optimization. Each weekly decision uses weather information available at that time (observations, short-range and long-range forecasts) and is re-optimized as conditions evolve, directly overcoming the fundamental foresight limitation of prior simulation-grounded advisory systems.
    \item The key component of Agri-SCHOLAR is a novel LLM-bound optimization mechanism that assigns complementary roles to language agents and numerical optimization. LLMs define agronomically feasible action bounds, while the optimizer searches within those bounds using APSIM as an evaluation oracle. This jointly optimizes productivity, profitability, resource use efficiency, and sustainability, as well as prevents any gaming of the simulator by the optimizer. This jointly optimizes productivity, profitability, resource use efficiency, and sustainability while preventing optimizer gaming. Backed by a dynamic RAG base, it doesn't require expert re-specification for new crops, regions, or seasons.
\item We provide a substantive empirical demonstration across 11 maize crop seasons in two contrasting irrigated and rainfed South Asian environments. The study demonstrates yield gains and shows improved economic and environmental outcomes. An ablation study showcases the value of agronomic LLM bounds, demonstrating that the proposed architecture drives practical, sustainable improvements.
\end{enumerate}

\section{Related Work}
Most LLM-based agricultural advisory systems are grounded in retrieved text. ShizishanGPT \cite{yang2024shizishangpt} and AgriGPT \cite{yang2025agrigpt} pair general-purpose LLMs with knowledge graphs and multi-channel retrieval. AgriRegion \cite{fanuel2025agriregion} and AgroAskAI \cite{cantonjos2026agroaskai} improve localization through geospatial re-ranking and multi-agent governance, respectively, to reduce contextual hallucination. Even benchmarks such as AGXQA \cite{kpodo2024agxqa}, AgriBench \cite{zhou2024agribench}, AgMMU \cite{gauba2025agmmu}, and AgEval \cite{arshad2024ageval} evaluate these systems on question-answering accuracy and multimodal reasoning. None of them checks whether a recommendation is physiologically viable for the crop.
 
A smaller line of work grounds LLM outputs in process-based crop simulators. DSSAT-LM \cite{kpodo2026closing} takes a neurosymbolic view, using DSSAT as a deterministic symbolic backbone while the LLM interprets questions and generates answers. Agri-SAGE \cite{balasubramaniam2026agri} couples multi-agent LLM reasoning with APSIM \cite{mccown1996apsim} and compares Plan-and-Solve, Tree of Thoughts, and Reflexion as reasoning strategies over a 10-year retrospective in a single region, substantially improving physiological grounding over text-only advisory.

Agri-SCHOLAR builds on Agri-SAGE but differs in three ways. First, Agri-SAGE produces one season-long plan and refines it in APSIM using the observed weather of the entire season, which is unknown when the plan is made, so it cannot be deployed before or during a real season; Agri-SCHOLAR decides each week using only the information available on that date and re-optimizes as new observations arrive. Second, in Agri-SAGE the LLM proposes the management actions themselves, so the continuous space of doses and dates is explored through a handful of proposals; in Agri-SCHOLAR the LLM only sets agronomically admissible bounds, and a numerical optimizer evaluates many candidates within them in APSIM. Third, Agri-SAGE maximizes grain yield alone, whereas Agri-SCHOLAR maximizes net profit, accounting for production costs and monetized environmental penalties. Beyond these design differences, we evaluate Agri-SCHOLAR in two contrasting water regimes, irrigated and rainfed, rather than one.
 
Another body of work uses crop simulators as environments for learning management policies. gym-DSSAT \cite{gautron2022gym} wraps DSSAT in a standard Reinforcement Learning interface, and later work trains nitrogen and irrigation policies on top of it \cite{wu2022optimizing, tao2022optimizing}. \citet{wu2024new} verbalize the simulator’s state variables as text and train a language-model-based policy with deep Q-learning. \citet{wang2026learning} show that policies trained under normal conditions must be retrained for extreme weather, highlighting the value of adapting to in-season conditions rather than committing to a fixed season-long policy. These systems output numeric fertilization or irrigation schedules rather than explainable, farmer-facing advisories.
 
The unobserved remainder of a season cannot be reliably forecast either. Even the strongest machine-learned weather models, including GraphCast \cite{lam2022graphcast}, Pangu-Weather \cite{bi2023accurate}, GenCast \cite{price2025probabilistic}, and Aurora \cite{bodnar2025foundation}, are skillful only for roughly two weeks, well short of the 90-150 days of a crop cycle. Operational advisory services have adapted to this ceiling by issuing short-horizon guidance and revising it as the season unfolds. India's Gramin Krishi Mausam Sewa issues district-level advisories twice a week from roughly 130 Agro-Met Field Units, combining observed weather, short-range forecasts, and crop stage \cite{singh2023decision}. China's  meteorological services follows a similar rolling cadence, nowcasting precipitation at 0-3 days and extending guidance to 30 days through provincial and district agrometeorological service centers. Vietnam's agroclimatic bulletin program covering 13 provinces and roughly 290,000 farmers, issues comparable short-horizon, crop-stage-specific advisories \cite{le2025vietnamese}. These services are expert-driven and issued at the level of an administrative region.
 
Agri-SCHOLAR combines these threads. Like retrieval-grounded systems, it produces natural-language, farmer-facing advisories, but grounds them in a crop simulator. Like Agri-SAGE and DSSAT-LM, it uses a process-based simulator. However, it evaluates candidate actions on a trajectory built from observed weather, a short-range forecast, and an analog-year extrapolation, rather than assuming full-season foresight. Agri-SCHOLAR also optimizes against a simulator while restricting the search to LLM-proposed, agronomically feasible bounds, and scores candidates using a multi-objective reward that spans yield, resource-use efficiency, and economic feasibility, ensuring the outputs remain interpretable. Finally, it reissues advisories on a rolling cycle that mirrors how operational agro-meteorological units operate, while adding farm-level personalization and mechanistic validation that those services lack.

\begin{figure}[ht]
    \centering
    \includegraphics[width=\linewidth]{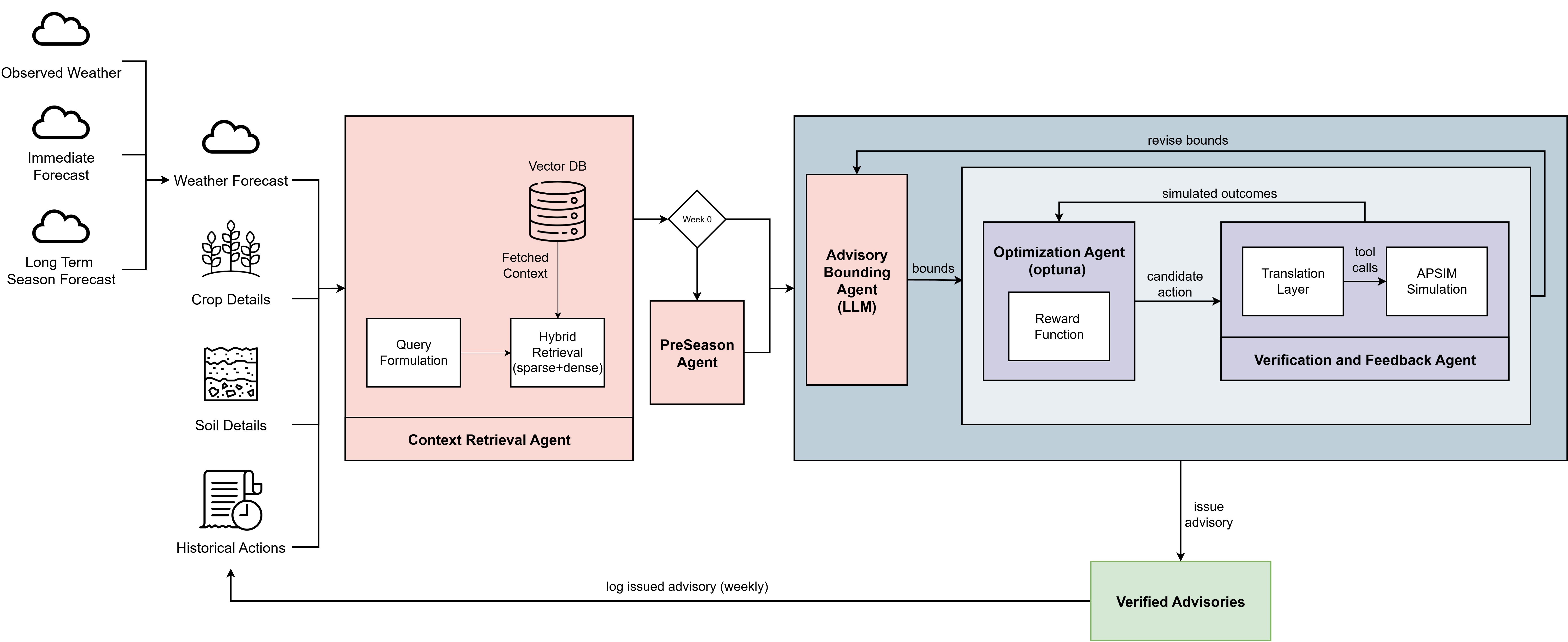}
    \caption{Agri-SCHOLAR architecture.}
    \label{fig:Agri-SCHOLAR-arch}
\end{figure}

\section{Methodology}

Figure~\ref{fig:Agri-SCHOLAR-arch} shows the architecture of Agri-SCHOLAR. The pipeline has five agents. (1) The Context Retrieval Agent (CRA), (2) PreSeason Agent (PSA), (3) Advisory Bounding Agent (ABA), (4) Optimization Agent (OA), and (5) Verification and Feedback Agent (VFA).

The PSA operates at the beginning of each cropping season to establish the initial seasonal context and constraints, while the remaining four agents jointly produce a short-term advisory each week using a constructed weather. Once the advisory has been optimized and verified, it passes through a cutoff layer, is issued to the farmer, and is logged as a historical action that conditions later cycles.

\subsection{Context Retrieval Agent (CRA)}
The CRA converts numerical weather, soil, and crop-stage features into a natural-language query. It then runs hybrid (sparse and dense) retrieval over a vector database of agronomic literature and regional guidance. Its output is a compact natural-language context bundle that the downstream agents reason over.

\subsection{PreSeason Agent (PSA)}\label{sec:psa}
The PSA sets the week-0 plan: the sowing date, mulch, and basal Di-Ammonium Phosphate (DAP) application. The sowing date is chosen by a rainfall-onset rule. The rule scans the period from June 1 to July 15 using a sliding three-day window and selects the first window with cumulative rainfall of at least 20~mm. The last day of that window is taken as the onset day, which serves as the sowing date (the agronomic planting day).

\subsection{Advisory Bounding Agent (ABA)}
Given the location details and the retrieved context, the ABA generates the advisory together with bounds for each management operation under consideration (e.g., plausible fertilizer amounts at the current crop stage). These bounds serve two purposes. First, they shrink the search space, because the OA searches only within the ABA's bounds rather than the full numerically feasible space. Second, they stop the OA from exploiting the simulator, for example: unrealistically large nitrogen applications that inflate simulated yield without being physiologically or practically plausible. Grounding the bounds in literature before APSIM is queried keeps the search within actions a farmer could realistically carry out.

\subsection{Optimization Agent (OA)}
\label{sec:optimizer}
At each weekly decision point, the OA searches within the ABA's bounds for the management action that maximizes expected economic return. Each candidate is evaluated by the VFA on the constructed weather file for the current cycle. This weather file contains only the information an advisory system would have at that point in the season, not the full observed record. The resulting APSIM simulation gives the projected crop response, from which the reward is computed.

Let $\pi = \{A_1, A_2, \ldots, A_K\}$ be a management policy, where $A_k$ is the set of fertilizer and irrigation decisions for week $k$ of a season with $K$ weekly decision points. The OA seeks the policy that maximizes cumulative profit:
\begin{equation}
\pi^{*} = \arg\max_{\pi} R(\pi),
\label{eq:profit-max}
\end{equation}
where $R(\pi)$ is the profit per hectare over the season under $\pi$, as simulated by APSIM.


Eq.~\ref{eq:profit-max} is solved each week $t$ using Optuna. At week $t$, all previously issued actions are fixed, and the OA jointly optimizes this week's urea and irrigation within the ABA's bounds together with a tentative nitrogen and irrigation schedules for the remaining weeks, constrained by the season structure set by the PSA (remaining splits, quantity, and minimum application). Each candidate is scored by a full APSIM run from sowing to harvest on the constructed weather $W_t$, using a Tree-structured Parzen Estimator with $25 + 4(K - t - 1)$ candidate evaluations per ABA iteration. Only the week-$t$ action is committed and issued; the tentative schedule is discarded and re-optimized the following week with updated observations.


Internally, fertilizer actions are expressed in kg N ha$^{-1}$ and denoted \emph{UreaN} (the quantity of elemental nitrogen applied), so UreaN is 100\% nitrogen by definition. Commercial urea, in contrast, contains only 46\% nitrogen by mass. So for farmer-facing advisories, each UreaN dose is therefore converted to commercial urea as follows
\begin{equation}
    m_{\mathrm{urea}} = \frac{m_{\mathrm{UreaN}}}{0.46},
    \label{eq:urea-conversion}
\end{equation}
For example, 46~kg~N\,ha$^{-1}$ of UreaN corresponds to 100~kg\,ha$^{-1}$ of commercial urea. All nitrogen constraints in the optimization (minimum dose, remaining splits, and season budget) are stated in UreaN units. Throughout this paper, we use UreaN for explanations and observations, and convert to commercial urea only in the farmer-facing advisories.

\subsubsection{Profit-Based Objective}
OA optimizes for farm profit based on yield, costs and environmental impact. Environmental impacts enter as monetary \emph{sustainability penalties}. The reward per hectare is
\begin{equation}
R = R_{\text{rev}} - C_{\text{prod}} - P_{\text{sust}},
\label{eq:reward}
\end{equation}
where $R_{\text{rev}}$ is gross grain revenue, $C_{\text{prod}}$ is the production cost, and $P_{\text{sust}}$ is the sustainability penalty. All monetary quantities are expressed per hectare in the local currency.

\paragraph{Revenue and production costs.}
Revenue is the simulated grain yield $Y$ (kg\,ha$^{-1}$) multiplied by the grain price. Production cost is the sum of fertilizer, irrigation, and labour costs, all of which follow directly from the actions chosen by the OA:
\begin{align}
R_{\text{rev}} &= p_g Y, \label{eq:revenue} \\
C_{\text{prod}} &= C_F + C_I + C_L, \label{eq:cprod} \\
C_F &= \sum_{i=1}^{n_F} A_i^F\, p_i^F, \label{eq:cf} \\
C_I &= \sum_{j=1}^{n_I} A_j^I\, p_w, \label{eq:ci} \\
C_L &= (n_F + n_I)\, p_L. \label{eq:cl}
\end{align}
Here $C_F$, $C_I$, and $C_L$ are the fertilizer, irrigation, and labour costs. $n_F$ and $n_I$ are the numbers of fertilizer and irrigation operations in the season. $A_i^F$ is the amount of fertilizer applied in the $i$-th fertilizer operation (kg\,ha$^{-1}$), and $A_j^I$ is the amount of water applied in the $j$-th irrigation operation (mm). The unit prices are $p_g$ for grain (per kg), $p_i^F$ for fertilizer in operation $i$ (per kg), $p_w$ for irrigation water (per mm), and $p_L$ for labour (per operation).

\paragraph{Sustainability penalties.}
The sustainability penalty $P_{\text{sust}}$ combines impacts that occur during the season ($P_{\text{season}}$) with impacts that appear after it ($P_{\text{post}}$):
\begin{align}
P_{\text{sust}} &= P_{\text{season}} + P_{\text{post}}, \label{eq:psust} \\
P_{\text{season}} &= S_{wl} I_{wl} + S_{no} I_{no}, \label{eq:pseason} \\
P_{\text{post}} &= S_{dd} I_{dd} + S_{nl} I_{nl} + S_{cl} I_{cl}. \label{eq:ppost}
\end{align}
Each term multiplies an environmental \emph{indicator} $I_{\cdot}$, computed from APSIM outputs, by a \emph{shadow cost} $S_{\cdot}$, the monetary cost assigned to one unit of that impact. The subscripts denote waterlogging ($wl$), nitrogen oversupply ($no$), deep drainage ($dd$), nitrogen leaching ($nl$), and soil organic carbon loss ($cl$). Waterlogging and nitrogen oversupply are penalized in-season because they directly harm the crop and waste inputs. The other three are penalized after the season because they represent water and nutrient losses beyond the root zone and the depletion of long-term soil fertility.

With $T$ the number of days in the simulated season, the indicators are
\begin{align}
I_{wl} &= \frac{1}{T}\sum_{t=1}^{T} \max\!\left(0,\ \frac{SW_t - DUL_t}{SAT_t - DUL_t}\right), \label{eq:iw} \\
I_{no} &= \frac{1}{T}\sum_{t=1}^{T} \max\!\left(0,\ N_{\text{applied},t} - N_{\text{demand},t}\right), \label{eq:in} \\
I_{dd} &= D_{\text{cml}}, \label{eq:id} \\
I_{nl} &= L_{NO_3} + L_{NH_4}, \label{eq:il} \\
I_{cl} &= \max\!\left(0,\ SOC_0 - SOC_H\right). \label{eq:ic}
\end{align}
Economic quantities come from the OA's selected actions and external prices, whereas productivity and environmental indicators come exclusively from APSIM. Table~\ref{tab:provenance} lists the details of each quantity.

\begin{table}[ht]
\centering
\small
\caption{Quantities used in the reward and their sources.}
\label{tab:provenance}
\begin{tabularx}{\linewidth}{>{\raggedright\arraybackslash}p{0.17\linewidth}>{\raggedright\arraybackslash}X>{\raggedright\arraybackslash}X}
\toprule
\textbf{Symbol} & \textbf{Meaning} & \textbf{Source} \\
\midrule
$Y$ & Grain yield & APSIM: \texttt{[Maize].Grain.Wt} \\
$SW_t$, $DUL_t$, $SAT_t$ & Soil water, drained upper limit, and saturation of the profile on day $t$ & APSIM: \texttt{[Soil].Physical.}\allowbreak\texttt{[SWmm/DULmm/SATmm]} \\
$N_{\text{demand}}$ & Crop nitrogen demand, summed over leaf, stem, root, and grain organs & APSIM: sum of \texttt{NDemand.Total} over the organs \\
$D_{\text{cml}}$ & Cumulative deep drainage & APSIM: \texttt{[WaterBalance].Drainage} \\
$L_{NO_3}$, $L_{NH_4}$ & Cumulative nitrate and ammonium leaching & APSIM: \texttt{[WaterBalance].}\allowbreak\texttt{[LeachNO3/LeachNH4]} \\
$SOC_0$, $SOC_H$ & Soil organic carbon (SOC) at the start and end of the horizon & APSIM: \texttt{[Soil].Organic.Carbon} \\
$C_F$, $C_I$, $C_L$ & Fertilizer, irrigation, and labour costs & OA: computed from applied inputs, number of operations, and unit prices \\
$p_g$, $p_i^F$, $p_w$, $p_L$ & Unit prices of grain, fertilizer, water, and labour & External: regional or market prices \\
\bottomrule
\end{tabularx}
\end{table}

\subsection{Verification and Feedback Agent (VFA)}
The VFA translates each candidate action into APSIM tool calls and runs the simulation on the constructed weather file for the current cycle. It returns the resulting metrics (grain yield, soil water status, nitrogen balance, drainage, and leaching) to the OA, which computes the reward. It also returns them to the ABA, so that bounds can be tightened or relaxed if they repeatedly yield infeasible or dominated candidates.

\subsection{Iterative Loops}
Agri-SCHOLAR runs three nested loops at different frequencies.
\begin{itemize}
    \item \textbf{Inner loop (OA-VFA).} The OA proposes a candidate, the VFA scores it in APSIM, and the OA updates its search.
    \item \textbf{Middle loop (ABA-OA/VFA).} When the inner loop converges, or fails to find a satisfactory candidate, the ABA uses the VFA's feedback to revise the bounds through self-reflection, and the inner loop is repeated within the revised bounds.
    \item \textbf{Outer loop (weekly).} The full pipeline reruns from scratch each week with the past week's observed weather, a new short-range forecast, and an updated analog tail, producing a new advisory.
\end{itemize}

\subsection{Weather Array Construction}
\label{sec:weather-array}
To emulate the information available to an operational advisory system, we construct the weather as follows: weather up to current date, a noisy seven-day forecast, and an analog-based estimate for the remainder of the season.

Let the crop be sown on day $S$, with season length $D$ days. At week $t \geq 0$ after sowing, the weather file supplied to APSIM is
\begin{equation}
W_t = C_{\mathrm{pre}} \,\Vert\, O_{1:7t} \,\Vert\, F_{7t+1:7t+7} \,\Vert\, \hat{A}_{7t+8:D},
\label{eq:weather-array}
\end{equation}
where
\begin{itemize}
    \item $C_{\mathrm{pre}}$ is observed weather from January~1 to day $S-1$;
    \item $O_{1:7t}$ is observed weather from sowing through the current day;
    \item $F_{7t+1:7t+7}$ is a synthetic seven-day forecast obtained by perturbing the true weather;
    \item $\hat{A}_{7t+8:D}$ is the analog tail, an observed weather sequence from the historically most similar season.
\end{itemize}

The segment $F$ mimics a real seven-day forecast by adding errors to the observed weather that scale with lead time. All other segments rely strictly on data available on the decision date, preventing the model from seeing any seasonal information beyond the seven-day horizon.

\subsubsection{Synthetic Seven Day Forecast}
The forecast segment $F_{7t+1:7t+7}$ is generated by perturbing the true weather with stochastic errors that grow with lead time. For forecast lead time $\tau \in \{1, \ldots, 7\}$ days, precipitation is simulated first because it conditions the temperature and radiation errors.

\paragraph{Precipitation.}
Precipitation follows the wet/dry occurrence framework of Richardson's stochastic weather generator \cite{richardson1981stochastic}. First, the day's wet/dry state is flipped with probability
\begin{equation}
P_{\mathrm{flip}}(\tau) = 0.05\,\tau^{0.7}.
\end{equation}
If the resulting day is dry, precipitation is set to zero. Otherwise, the amount is sampled from a Gamma distribution, following the Weather Generator (WGEN) \cite{richardson1984wgen}:
\begin{equation}
\mathrm{P}_f(\tau) \sim \mathrm{Gamma}\!\left(\mathrm{shape}=\frac{1}{\sigma_p^2 \tau^{b}},\ \mathrm{scale}=\mathrm{P}_{\mathrm{true}}\,\sigma_p^2 \tau^{b}\right),
\end{equation}
where $\mathrm{P}_f$ and $\mathrm{P}_{\mathrm{true}}$ denote forecast and true precipitation. This parameterization preserves the expected precipitation, $\mathbb{E}[\mathrm{P}_f(\tau)] = \mathrm{P}_{\mathrm{true}}$, while letting the relative uncertainty grow with lead time.

To propagate precipitation uncertainty to the other variables, we define the outcome $\delta_{\mathrm{precip}}(\tau) \in \{-1, 0, +1\}$. Here $+1$ denotes a false alarm or overprediction, $-1$ a missed or underpredicted event, and $0$ a preserved precipitation state.

\paragraph{Temperature.}
Temperature errors are additive Gaussian with lead-time-dependent variance and a small mean shift conditioned on $\delta_{\mathrm{precip}}(\tau)$:
\begin{equation}
\begin{gathered}
T_f(\tau) = T_{\mathrm{true}}(\tau) - \delta_{\mathrm{precip}}(\tau)\,\mu_T(\tau) + \epsilon_T(\tau), \\
\epsilon_T(\tau) \sim \mathcal{N}\!\left(0,\ \sigma_T^2 \tau^{2\beta}\right).
\end{gathered}
\end{equation}
The parameters $\sigma_T$ and $\beta$ are estimated from the empirical validation of the U-Cast probabilistic forecaster \cite{cachay2026u}. Converting the reported day-1 and day-7 temperature  Continuous Ranked Probability Score into equivalent standard deviations gives a base uncertainty of $\sigma_T \approx 0.93^{\circ}$C and a lead-time growth exponent of $\beta \approx 0.36$. The mean shift $\mu_T(\tau)$ (approximately $0.3$-$0.8^{\circ}$C) is calibrated in this study and applied only when the precipitation state changes, reflecting the effect of cloud cover on surface temperature.

\paragraph{Solar radiation.}
Solar radiation is perturbed in clear-sky-index space rather than directly in irradiance. The clear-sky index is
\begin{equation}
\kappa = \frac{\mathrm{Rad}_{\mathrm{true}}}{\mathrm{Rad}_{\mathrm{cs}}},
\end{equation}
where $\mathrm{Rad}_{\mathrm{cs}}$ is the clear-sky irradiance from the Ineichen-Perez model \cite{ineichen2002new}. Working in this normalized space removes predictable seasonal and astronomical variation and keeps perturbed values physically bounded. Gaussian forecast error is then applied:
\begin{gather}
\kappa_f(\tau) = \mathrm{clip} \!\Big(\kappa_{\mathrm{true}}(\tau) - \delta_{\mathrm{precip}}(\tau)\,\mu_\kappa(\tau) + \epsilon_\kappa(\tau),\ 0,\ 1.1\Big), \\
\epsilon_\kappa(\tau) \sim \mathcal{N}\!\left(0,\ \sigma_\kappa^2 \tau^{2\beta}\right),
\end{gather}
and the forecast radiation is recovered as
\begin{equation}
\mathrm{Rad}_f(\tau) = \kappa_f(\tau)\,\mathrm{Rad}_{\mathrm{cs}}(\tau).
\end{equation}
As with temperature, the shift $\mu_\kappa(\tau)$ is calibrated for this study. It lowers the clear-sky index on forecast wet days and raises it on forecast dry days, preserving the physical link between cloud cover, precipitation, and incoming radiation.

\subsubsection{Analog Tail Selection}
The weather beyond the 7-day forecast is filled in with observed weather from a historical year that resembles the current season. The analog is selected using only information available up to the decision date. For a target season in year $y$, the candidate pool is $\mathcal{H}_y = \{1980, \ldots, y-1\}$: the target year and all later years are excluded, so neither the realized weather of the season nor any subsequent record can enter the analog tail. For example, an advisory for the 2015 season compares the weather observed up to its decision date with the same calendar window in each year from 1980 to 2014.

At each cycle $t$, we describe the current season with a climate descriptor
\begin{equation}
X_t = \left[\mathrm{ONI},\ \mathrm{P},\ \Delta T_{\max}\right],
\end{equation}
which captures the large-scale climate state and the seasonal buildup before sowing. Here ONI is the Oceanic Niño Index, the running three-month mean of sea-surface-temperature anomalies. $\mathrm{P}$ and $\Delta T_{\max}$ are the cumulative precipitation and the maximum-temperature anomaly over the previous weeks.

Because these features have different units and scales, we standardize them. Let $H$ be the historical feature matrix, with row $i$ a year in $\mathcal{H}_y$ and column $j$ a feature, and let $\mu_j$ and $\sigma_j$ be the mean and standard deviation of feature $j$ over $\mathcal{H}_y$. Then
\begin{equation}
Z_{i,j} = \frac{H_{i,j} - \mu_j}{\sigma_j}, \qquad z_{t,j} = \frac{X_{t,j} - \mu_j}{\sigma_j}.
\end{equation}
Each feature carries a fixed agronomic importance weight, with $\boldsymbol{\omega} = [\omega_1, \ldots, \omega_k]$ and $\sum_{j=1}^{k} \omega_j = 1$. The dissimilarity between the current season and historical year $i$ is the weighted Euclidean distance
\begin{equation}
d_i = \sqrt{\sum_{j=1}^{k} \omega_j \left(Z_{i,j} - z_{t,j}\right)^2}.
\end{equation}
We select the year $i^{*} = \arg\min_{i \in \mathcal{H}_y} d_i$.

\section{Experimental Results}
\label{sec:experiments}
 
This section begins with the experimental setup (Section~\ref{sec:setup}). It then compares Agri-SCHOLAR against the regional package of practices on yield (Section~\ref{sec:yield}), profit (Section~\ref{sec:profit}), and sustainability (Section~\ref{sec:sustain}). Next, it examines a single season in detail as a case study (Section~\ref{sec:case}) and closes with an ablation study of the Advisory Bounding Agent (Section~\ref{sec:aba_section}).
 
\subsection{Experimental Setup}
\label{sec:setup}
 
We evaluate Agri-SCHOLAR for maize in two regions of South Asia that differ in their water regimes: one irrigated and one rainfed. For each region, we simulate 11 consecutive kharif seasons (2015-2025). Although we study maize, the framework is crop-agnostic: it can be applied to any crop the simulator supports, and another process-based simulator can replace APSIM. All language agents use \textit{google/gemma-4-E4B-it}.
 
\subsubsection{Data}
Daily weather (rainfall, minimum and maximum temperature, solar radiation) is taken from ERA5 for 1980-2025 at the coordinates of each study site and serves as the ``observed'' weather.
The full record is used only to evaluate each season; advisories use only data available up to the decision date (Section~\ref{sec:weather-array}). Local agricultural experts provided soil profiles for both regions. The knowledge base used by the CRA consists of agronomic literature, institutional reports, and the regional package of practices for maize. The expert-curated costs for grain, fertilizer, water, and labor, along with sustainability shadow costs and references, are detailed in Appendix~\ref{app:params}.
 
\subsubsection{Baseline}
Our baseline is the regional package of practices (PoP), the main channel through which agronomic guidance reaches farmers. The PoP is a fixed calendar of fertilizer and irrigation operations, and we simulate it in APSIM under the actual weather of each season, representing a farmer who follows it diligently. Both methods sow on the same day, as given by the rainfall-onset rule of the PSA (Section~\ref{sec:psa}): from June~1 to July~15, we slide a three-day window and take the last day of the first window with at least 20~mm of cumulative rainfall. Differences in outcome, therefore, come from in-season management rather than from the sowing date.
 
\subsubsection{Evaluation Protocol and Metrics}
In each season, Agri-SCHOLAR issues one advisory per week, from sowing (week~0) until the crop reaches physiological maturity. Each advisory is chosen using only the constructed weather $W_t$ of that week (Eq.~\ref{eq:weather-array}), never the future observed weather. The advisories issued over the season form the executed management plan, which is fixed once issued. The full season is then replayed in APSIM under the true weather to obtain the realized outcome. The PoP is evaluated on the same true weather, so both methods are scored on identical seasons, and Agri-SCHOLAR's results reflect decisions made without knowledge of the weather beyond the seven-day forecast.
 
We report grain yield, net profit $R$ (Eq.~\ref{eq:reward}), and its components: revenue, production cost, and sustainability penalty. Monetary values are converted from local currency to US dollars at purchasing power parity (Appendix~\ref{app:params}). Means are taken over the 11 seasons of each region, and improvements are stated relative to the PoP.
 
\subsection{Yield}
\label{sec:yield}
 
\begin{figure}[ht]
    \centering
    \begin{minipage}[t]{0.495\linewidth}
        \centering
        \includegraphics[width=\linewidth]{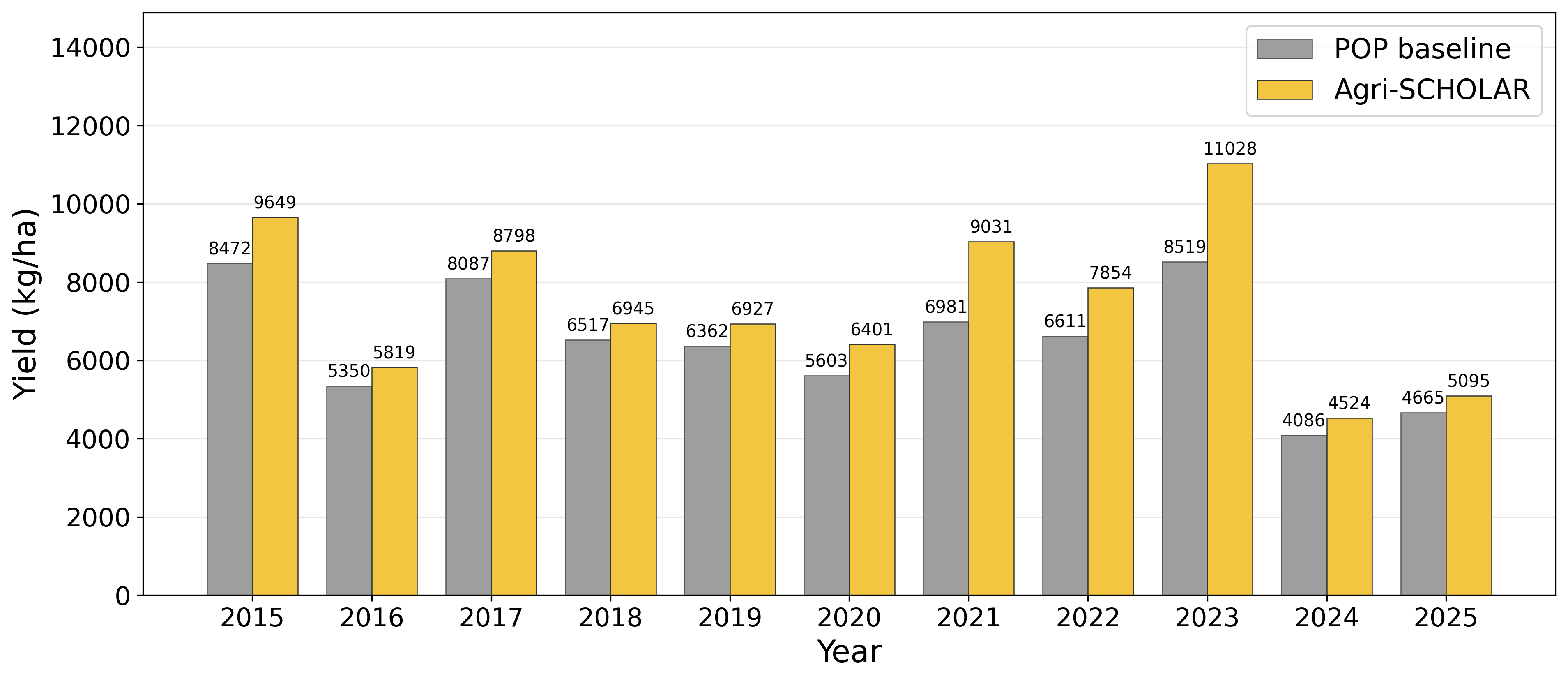}\\
        \small (a) Irrigated region
    \end{minipage}\hfill
    \begin{minipage}[t]{0.495\linewidth}
        \centering
        \includegraphics[width=\linewidth]{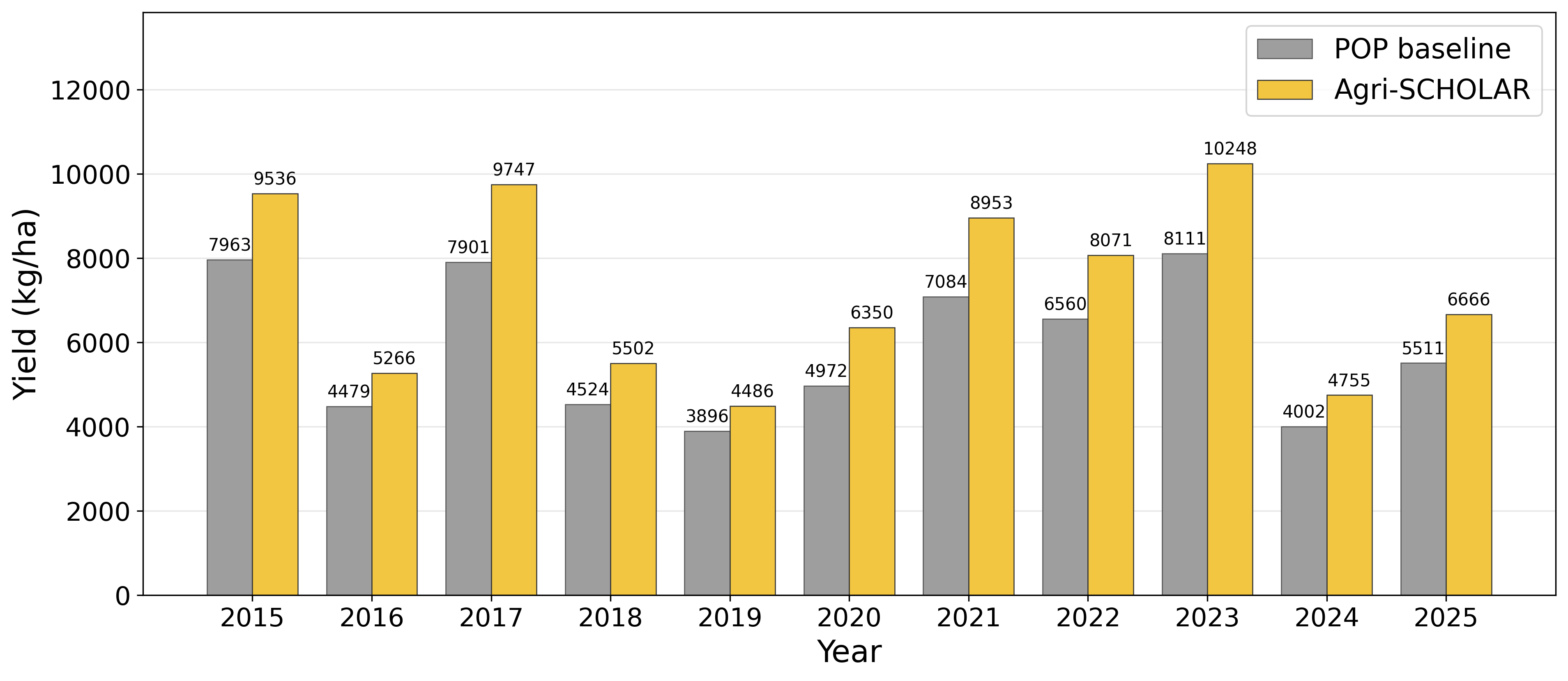}\\
        \small (b) Rainfed region
    \end{minipage}
   \hfill
    \caption{Grain yield per season under PoP baseline and Agri-SCHOLAR.}
    \label{fig:yield}
\end{figure}
 
Figure~\ref{fig:yield} shows grain yield per season in the irrigated and rainfed regions. Agri-SCHOLAR outperforms the PoP in every season of the study in both regions. In the irrigated region, it raises mean yield by 15.18\%, with per-season gains ranging from 6.57\% to 29.45\%. In the rainfed region, the mean improvement is 22.43\%, with per-season gains ranging from 15.14\% to 27.72\%.
 
The size of the gain depends on how the season's rainfall is distributed. In the irrigated region, the gain is largest in 2023 (+2,509~kg\,ha$^{-1}$, +29.45\%), a season with a long mid-season dry spell. The PoP's fixed calendar cannot respond to such a spell, whereas Agri-SCHOLAR detects it from the observed weather and the short-range forecast and irrigates when soil moisture falls (Section~\ref{sec:case}). The gain is smallest in 2018 (+428~kg\,ha$^{-1}$, +6.57\%), a season with well-distributed rainfall, in which the PoP's calendar happens to suit the weather and leaves little room for improvement.
 
\begin{figure}[ht]
    \centering
    \includegraphics[width=0.7\linewidth]{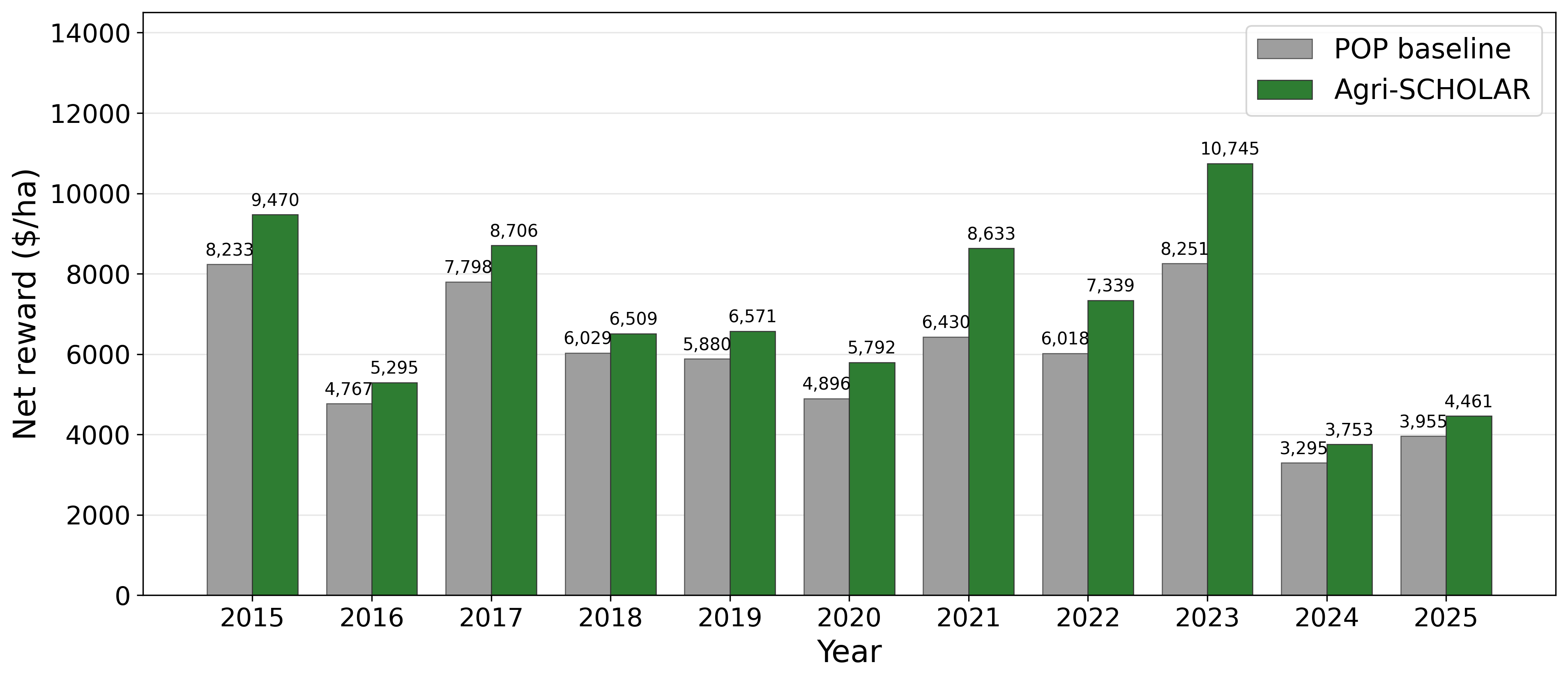}
    \caption{Net profit per season under PoP baseline and Agri-SCHOLAR (irrigated region).}
    \label{fig:revenue}
\end{figure}

\subsection{Profit and Resource Use}
\label{sec:profit}
 
Figure~\ref{fig:revenue} reports net profit $R$ per season in the irrigated region, that is, grain revenue minus production cost and sustainability penalty (Eq.~\ref{eq:reward}). Agri-SCHOLAR earns a higher net profit than the PoP in every season in both regions, and it increases mean net profit by 17.2\% in the irrigated region and by 22.8\% in the rainfed region. Table~\ref{tab:main} summarizes mean yield, profit, production cost, and sustainability penalty.
 
Table~\ref{tab:main} also shows where the profit gain comes from. Agri-SCHOLAR spends more on inputs than the PoP: its mean production cost is \$185.66~ha$^{-1}$ higher in the irrigated region and \$64.46~ha$^{-1}$ higher in the rainfed region. This extra spending is recovered several times over by the additional grain. At the grain price of \$1.04~kg$^{-1}$ (Table~\ref{tab:params}), the mean yield gain of 983~kg\,ha$^{-1}$ in the irrigated region is worth about \$1{,}023~ha$^{-1}$, and the sustainability penalty is \$228~ha$^{-1}$ lower. Together these account for the mean profit gain of \$1{,}066~ha$^{-1}$. In the rainfed region, the yield gain of 1{,}325~kg\,ha$^{-1}$ is worth about \$1{,}378~ha$^{-1}$, while the penalty is only \$8~ha$^{-1}$ lower, so almost the entire profit gain of \$1{,}322~ha$^{-1}$ comes from higher yield. Agri-SCHOLAR therefore does not raise profit by cutting inputs. It spends more on inputs when the resulting yield gain outweighs their cost, and in the irrigated region it also reduces environmental losses, as discussed next.

\begin{table}[ht]
\centering
\small
\setlength{\tabcolsep}{3pt}
\caption{Mean yield, net profit, production cost, and sustainability penalty over 11 seasons. Monetary values are in PPP-adjusted \$/ha. Best value per column and region in bold (highest for yield and profit, lowest for cost and penalty).}
\label{tab:main}
\begin{tabular}{@{}llrrrr@{}}
\toprule
Region & Method & Yield(kg/ha) & Profit(\$/ha)& \shortstack{Prod.\\cost} & \shortstack{Sust.\\penalty} \\
\midrule
Irrigated & PoP & 6,477.55 & 5,959.33 & \textbf{284.02} & 495.33 \\
 & Agri-SCHOLAR & \textbf{7,461.00} & \textbf{7,025.09} & 469.68	 &\textbf{267.05}	 \\
\midrule
Rainfed & PoP & 5,909.36 & 5715.15 &\textbf{ 149.63} & 282.81 \\
 & Agri-SCHOLAR & \textbf{7,234.55} & \textbf{7,037.19} & 214.09 & \textbf{274.95}\\
\bottomrule
\end{tabular}
\end{table}

\begin{figure}[ht]
    \centering
    \begin{minipage}[t]{0.495\linewidth}
        \centering
        \includegraphics[width=\linewidth]{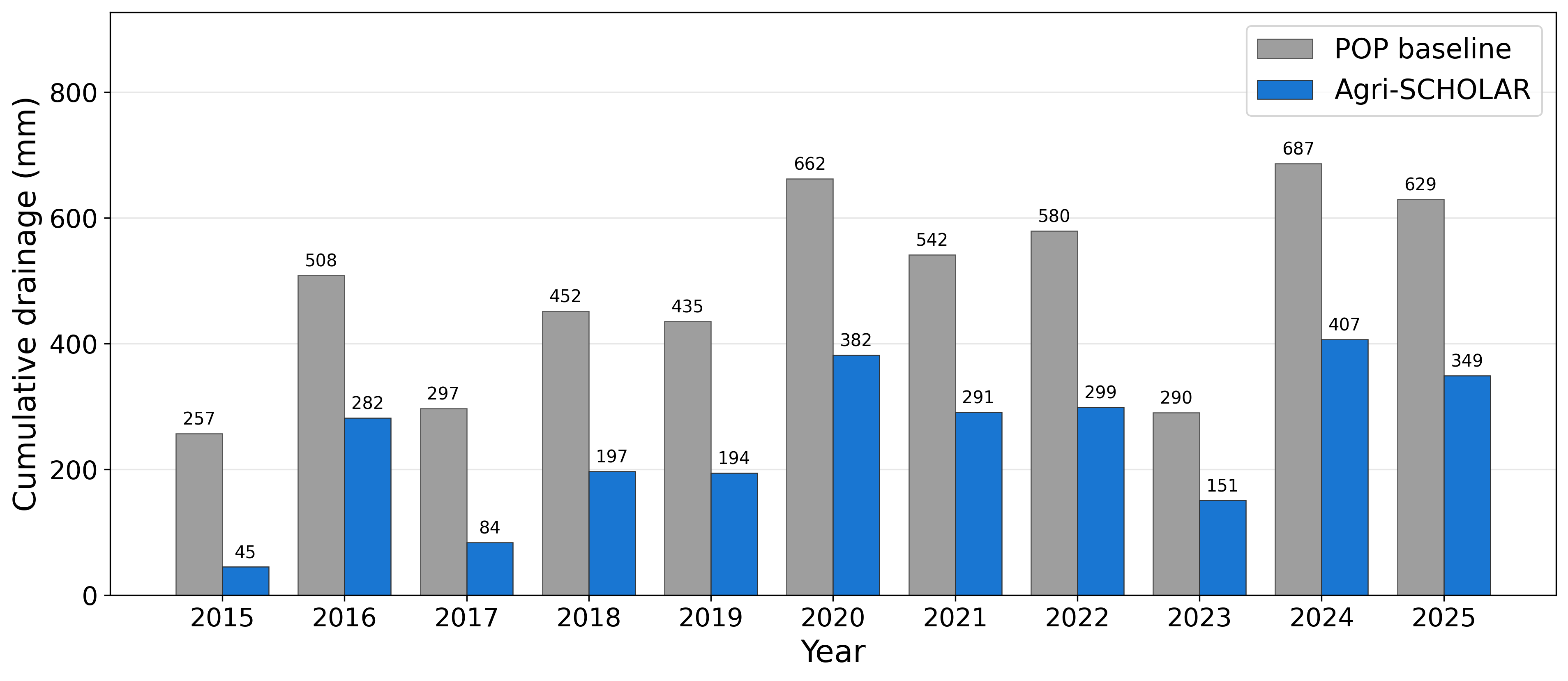}\\
        \small (a) Cumulative Deep Drainage
    \end{minipage}\hfill
    \begin{minipage}[t]{0.495\linewidth}
        \centering
        \includegraphics[width=\linewidth]{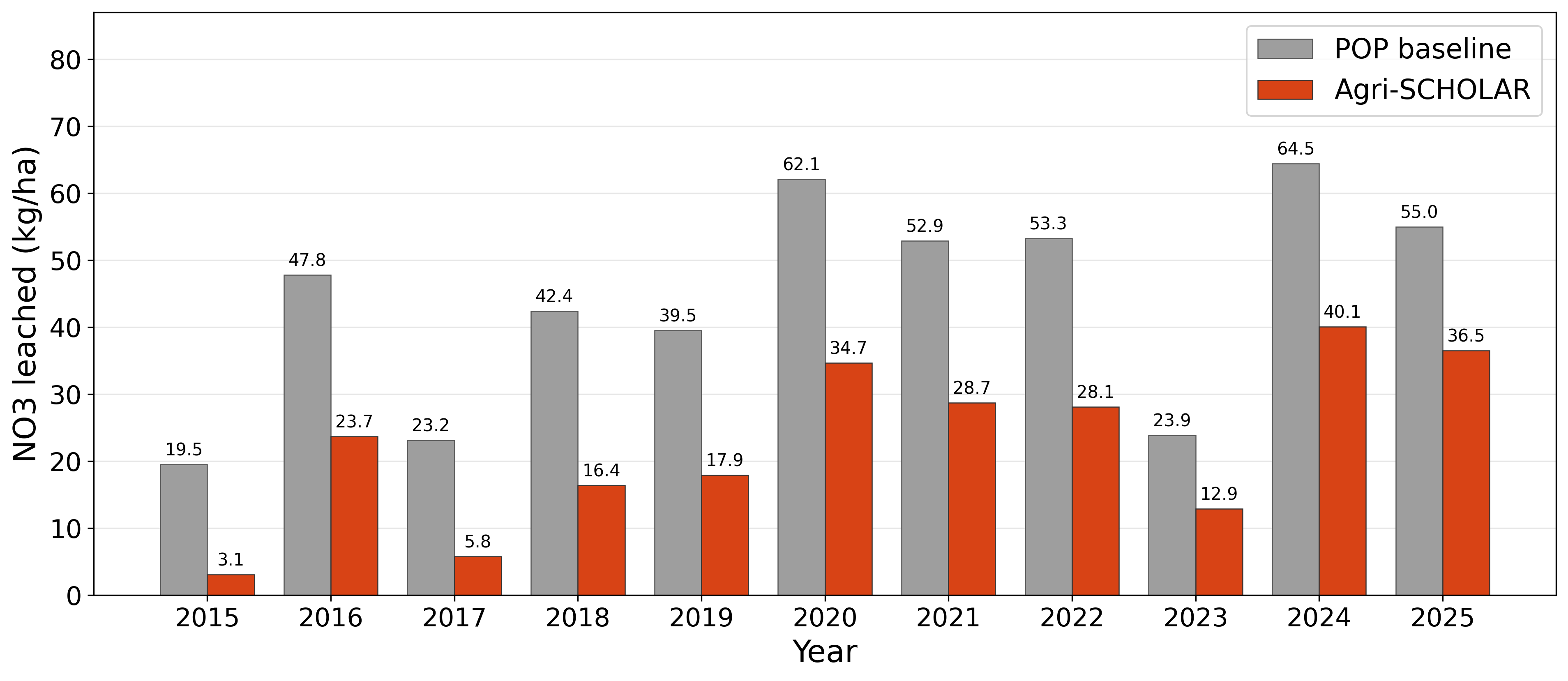}\\
        \small (b) Cumulative Nitrate Leaching
    \end{minipage}
   \hfill
    \caption{Drainage and Leaching per season (irrigated region).}
    \label{fig:sustain}
\end{figure}
 
\subsection{Sustainability}
\label{sec:sustain}
 
Figure~\ref{fig:sustain} compares cumulative deep drainage and nitrate leaching per season in the irrigated region. Deep drainage is the water that moves below the root zone and is lost to the crop, and nitrate leaching is the nitrogen carried below the root zone with it. Agri-SCHOLAR reduces mean drainage by 52.8\% and mean NO$_3$ leaching by 52.7\%. The reduction holds in every season, ranging from 40.73\% to 82.49\% for drainage and from 33.64\% to 84.13\% for leaching, and it lowers the mean sustainability penalty by 46.1\% (Table~\ref{tab:main}).
 
These reductions follow from the structure of the reward. Drainage and leaching carry shadow costs (Eq.~\ref{eq:ppost}), so irrigating soil that is already wet is penalized twice, through the cost of the water and through the drainage it causes. The optimizer therefore irrigates only when soil moisture is low. Similarly, nitrogen applied before the crop can take it up remains in the soil and is exposed to leaching, so the optimizer times UreaN doses to crop demand. The PoP irrigates and fertilizes on a fixed calendar and has no such feedback. The case study in Section~\ref{sec:case} shows both effects.
 
In the rainfed region, the mean penalty falls by only 2.8\% (Table~\ref{tab:main}). A likely reason is that, without irrigation, drainage in this region is driven mainly by rainfall, which management cannot change, so there is less room for improvement.

\subsection{Case Study}
\label{sec:case}
\begin{figure}[ht]
    \centering
    \includegraphics[width=\linewidth]{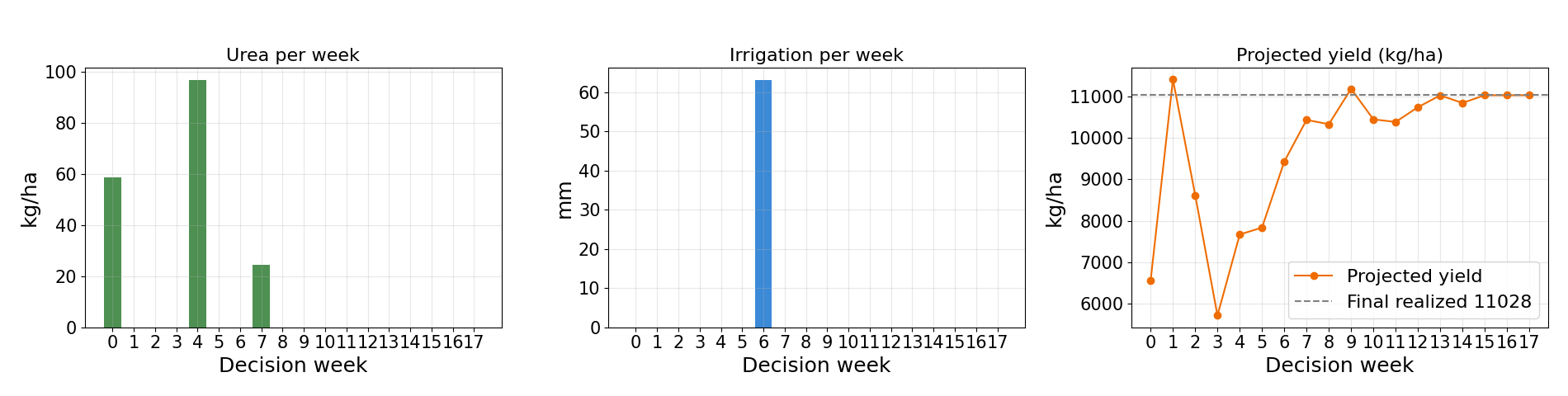}
    \caption{Case study for irrigated region, 2023 season. (a) UreaN applied per week (b) Irrigation per week (c) Projected Yield}
    \label{fig:case}
\end{figure}

To show how the rolling cycle behaves, we follow the 2023 kharif season in the irrigated region, chosen because it showed the largest gain over the PoP ($+2{,}509$~kg\,ha$^{-1}$, $+29.45\%$). Figure~\ref{fig:case} provides week-by-week information on Fertiliser (UreaN, DAP) and irrigation applied per week. It also shows how the projected yield changes over the season, and Listing~\ref{lst:advisory} presents selected natural-language advisories issued by the pipeline. The complete week-by-week advisories are provided in Appendix~\ref{app:case}.

\begin{figure*}[!t]
\centering
\begin{minipage}[t]{0.49\textwidth}
\begin{advisory}
# PRE-SOWING (land preparation)
sow:     04 Jul 2023 (monsoon set in; soil moist enough
         for germination)
basal:   200 kg/ha DAP (4 bags of 50 kg) in the
         seed furrow at sowing
mulch:   26 kg/ha residue/straw between rows to
         save soil moisture
tillage: light/reduced tillage (20% surface disturbed)

# WEEK 0 | 04 Jul | 0 Days After Sowing (DAS)
stage:   sowing / germination
weather: rain since 1 Jun = 88 mm; longest dry spell = 10 d
action:  apply 128.3 kg/ha of urea (2.85 bags of 45 kg)
         as side-dressing near plant rows on moist
         soil; cover with soil
why:     starter urea with basal DAP supports early root
         and leaf growth; model check: no water or N
         stress expected under this plan

# WEEK 4 | 01 Aug | 28 DAS
stage:   seedling to early vegetative
weather: rain since 1 Jun = 332 mm (42 mm past week);
         longest dry spell = 10 d
action:  apply 210.9 kg/ha of urea (4.69 bags of 45 kg)
         as side-dressing near plant rows on moist
         soil; cover with soil
why:     crop is entering its fastest growth phase and
         takes up most of its N now; top-dressing avoids
         leaf yellowing; model check: no water or N
         stress expected under this plan
\end{advisory}
\end{minipage}\hfill
\begin{minipage}[t]{0.49\textwidth}
\begin{advisory}
# WEEK 6 | 15 Aug | 42 DAS
stage:   knee-high, rapid vegetative growth
weather: rain since 1 Jun = 351 mm (7 mm past week);
         longest dry spell = 11 d
action:  one irrigation of 63 mm (631,000 L/ha;
         roughly one furrow irrigation)
why:     soil moisture is running low for the current
         crop stage (dry spells of up to 11 d this season);
         model check: minor stress remains (water 5.4 d,
         N 0.0 d)

# WEEK 8 | 29 Aug | 56 DAS
stage:   tasselling and silking (most water-sensitive)
weather: rain since 1 Jun = 362 mm (5 mm past week);
         longest dry spell = 23 d
action:  no fertiliser or irrigation this week; continue
         routine scouting (weeds, fall armyworm, leaf blight)
why:     only 5 mm rain last week, but soil still holds
         enough moisture; week-7 urea still available,
         extra urea now would mostly leach; model check:
         minor stress remains (water 5.6 d, N 0.0 d)

# WEEK 16 | 24 Oct | 112 DAS
stage:   physiological maturity / harvest preparation
weather: rain since 1 Jun = 451 mm (0 mm past week);
         longest dry spell = 23 d
action:  no fertiliser or irrigation this week
why:     crop is maturing, so fertiliser or water will no
         longer raise yield; model check: minor stress
         remains (water 2.9 d, N 0.0 d)
\end{advisory}
\end{minipage}
\caption{Selected weekly advisories issued by Agri-SCHOLAR in the 2023 irrigated-region case study.}
\label{lst:advisory}
\end{figure*}

The pipeline recommended sowing on 4 July, once the monsoon had set in, with basal DAP, residue mulching, and reduced tillage. In week~0, it added a starter UreaN dose, and the crop-model check predicted no water or nitrogen stress. The early season was wet (332~mm by week~4, including 101~mm in week~3), so the advisories withheld irrigation and extra UreaN, since the soil was moist and added nitrogen would have leached. In week~4, as the crop entered rapid growth, it prescribed the main top-dress of 97~kg\,ha$^{-1}$ UreaN. The most consequential decision came in week~6, when only 7~mm of rain had fallen in the past week and the dry spell had reached 11 days: a single 63~mm irrigation halved the projected water stress. By week~8 (tasselling and silking) the dry spell had grown to 23 days, yet no further irrigation was issued because the soil still held enough moisture. Over the season, the pipeline applied 180~kg\,ha$^{-1}$ of UreaN in three splits and one irrigation, yielding 11{,}028~kg\,ha$^{-1}$ with 12.9~kg\,N\,ha$^{-1}$ leached.

In comparison, the Package of Practices (PoP) follows a fixed schedule: a basal dose of 75~kg P$_2$O$_5$, 40~kg K$_2$O, and 50~kg N\,ha$^{-1}$, then 50~kg N\,ha$^{-1}$ at weeks~4 and~8 (150~kg N\,ha$^{-1}$ in total), plus six irrigations at weeks~0, 2, 4, 6, 8, and 10. Being tied to the calendar, it cannot respond to the season as it unfolds. It ignores rainfall (332~mm had fallen by week~4, yet it still irrigates at weeks~0, 2, and 4), fixes total nitrogen regardless of crop need, which raises cost and leaching in wet seasons, and spreads irrigation evenly, whereas Agri-SCHOLAR concentrated its single irrigation at week~6, when soil moisture was actually low. As a result, the PoP uses five more irrigation events and about 77\% more nitrogen, yet yields 8{,}519~kg\,ha$^{-1}$, which is 2{,}509~kg\,ha$^{-1}$ less than Agri-SCHOLAR. A farmer may adjust the PoP from experience, but it remains a single predefined plan, whereas Agri-SCHOLAR adapts to the evolving crop, soil moisture, and weather on each farm and season.

\FloatBarrier
\subsection{Ablation: Advisory Bounding Agent}\label{sec:aba_section}
The central design choice of Agri-SCHOLAR is that the optimizer does not search the action space on its own: the ABA first uses the retrieved agronomic context to decide which actions are admissible at the current crop stage, and the OA searches only within those LLM-generated bounds. To test this choice, we keep the simulator, reward, optimizer, weather information, and season structure fixed, and replace the ABA's bounds in two ways.
\begin{itemize}
    \item \textbf{Loose bounds.} The OA searches a single fixed range for each action, identical in every week and every season: urea-N in $[0, \overline{N}]$~kg\, N\,ha$^{-1}$ per application and irrigation in $[0, \overline{I}]$~mm per week, where $\overline{N}$ and $\overline{I}$ are upper limits. These limits are hardcoded using the regional package of practices, so they rely on expert-curated, region-specific knowledge and naturally do not adapt to crop stage, soil conditions, or weather. This makes the pipeline hard to scale, since each new region or crop would require its own manually specified limits, much like the expert-guided advisories.

    \item \textbf{No bounds.} The upper limits are removed, so the OA may propose any non-negative amount of UreaN and irrigation in any week.
\end{itemize}
 
Table~\ref{tab:ablation} reports the mean of each metric over the 11 seasons. The number below each mean, after $\pm$, is the standard deviation of that metric across the 11 seasons. It measures how much the outcome varies from one season to the next, it is not an uncertainty over repeated runs. 
 
With loose bounds, mean yield rises by 2.1\%, profit by 2.5\%, and production cost falls by 14.6\%, but the sustainability penalty rises by 20.1\%. These gains come from actions that a farmer could not or should not carry out, such as doses too small to apply (2.8~kg\,ha$^{-1}$ of UreaN at week~8 of 2024) or a single large dose at planting (190~kg\,ha$^{-1}$ of UreaN in 2019) that the young crop cannot yet take up and that is therefore exposed to leaching. The optimizer produces these schedules because neither APSIM nor the reward encodes whether a dose suits the crop's growth stage or can be applied with standard equipment (Appendix~\ref{app:ablation-obs}).
 
With no bounds, performance collapses. Mean yield falls by 5.1\%, production cost rises about 9.5-fold and the sustainability penalty about 4.6-fold, and mean profit drops by 76\%, with negative profit in four of the 11 seasons. In 2019, the OA applied 6{,}062~kg\,ha$^{-1}$ of UreaN and about 6{,}000~mm of irrigation in week~0, for a projected reward of about \$6,421.10 ha$^{-1}$ on the constructed weather. Under the realized weather, 3{,}271~kg\,ha$^{-1}$ of nitrate leached, the sustainability penalty reached about \$4,778.5 ha$^{-1}$, and the season ended with a net loss of \$2{,}887~ha$^{-1}$ (Table~\ref{tab:ablation_reward}).
 
These results show that the bounds guard against simulator exploitation and keep recommendations agronomically practical. The ABA also addresses the scaling problem: at finer spatial and temporal resolution, hand-specifying bounds for every crop stage, farm, and season becomes infeasible even for experts, whereas the ABA generates them automatically from the retrieved context for each new setting.  
\begin{table}[ht]
\centering
\small
\setlength{\tabcolsep}{4pt}
\caption{ABA ablation: irrigated region, mean over 11 seasons.}
\label{tab:ablation}
\begin{tabular}{@{}lrrrr@{}}
\toprule
Variant & \shortstack[r]{Yield (kg/ha)} & \shortstack[r]{Profit (\$/ha)} & \shortstack[r]{Prod.\ cost } & \shortstack[r]{Sust.\ penalty} \\
\midrule
Agri-SCHOLAR & 7,461.00 & 7,025.09 & 469.68 & 267.05 \\
\midrule
Loose bounds & 7,614.82 & 7,199.86 & 401.24 & 320.73 \\
             & {\scriptsize $\pm$1,785.82} & {\scriptsize $\pm$1,823.07} & {\scriptsize $\pm$61.40} & {\scriptsize $\pm$107.33} \\
No bounds    & 7,082.64 & 1,682.69 & 4,451.40 & 1,234.11 \\
             & {\scriptsize $\pm$2,153.72} & {\scriptsize $\pm$3,117.25} & {\scriptsize $\pm$2,521.46} & {\scriptsize $\pm$1,503.77} \\
\bottomrule
\end{tabular}
\end{table}

\section{Conclusion and Future Work}

We introduced Agri-SCHOLAR, a rolling, in-season framework for farm-specific crop advisories. Prior simulation-grounded advisory systems need the whole season's weather in advance, which no forecasting system can provide. Agri-SCHOLAR instead rebuilds its weather trajectory every week from observations to date, a short-range forecast, and an analog-year tail. Each advisory comes from a closed loop in which LLM agents propose agronomically feasible bounds on irrigation and fertilizer actions. An optimizer then searches within those bounds, with APSIM as the evaluation oracle, to maximize a profit-based reward that includes sustainability penalties.

Over 11 maize seasons (2015-2025) in one irrigated and one rainfed region of South Asia, Agri-SCHOLAR beat the regional package of practices (PoP) in every season, even though the PoP was simulated with perfect knowledge of the true weather. Mean yield rose by 15.2\% in the irrigated region and 22.4\% in the rainfed region, and mean net profit rose by 17.2\% and 22.8\%. In the irrigated region, deep drainage and nitrate leaching each fell by about 50\%. The 2023 case study shows where the gains come from. The pipeline withheld irrigation and extra nitrogen during a wet early season and concentrated its single irrigation at the week when soil moisture was low. Finally, we showed that the bounds from the Advisory Bounding Agent keep the optimizer from exploiting the simulator with infeasible actions. Taken together, these results suggest that rolling, simulator-validated advisories can overcome the main limitations of expert-driven, region-level guidance without an accurate season-long forecast.

\paragraph{\textbf{Future work.}} We see several directions for extending this work. First, we will explore reinforcement learning as an alternative or complement to the bounded optimizer, while keeping the resulting advisories interpretable. Second, we will study sowing-date decisions, which our experiments suggest matter considerably but which require much better long-range weather prediction than is available today. Third, we will replace the emulated forecasts with operational numerical weather prediction products and test how sensitive the advisories are to forecast error. Fourth, we will calibrate and validate the simulator against field trials, and learn the shadow costs and reward weights. Finally, we plan prospective pilots with farmers, delivering advisories through channels they already use, such as SMS or voice in local languages, to measure adoption and real-world outcomes and to feed their feedback back into the system.

\section*{Acknowledgments}
The authors thank Prof. Soma Dhawala for his valuable suggestions and agronomic insights. The authors also thank Dr. Richard Magala for his helpful discussion regarding the apsimNGpy framework.

The authors acknowledge the use of various AI tools and assistants, such as GPT and Claude, to enhance the writing quality and provide summaries of source materials. However, the authors diligently verified all information provided by these AI tools to ensure accuracy and factual integrity.

\bibliographystyle{unsrtnat}
\bibliography{references}

\clearpage
\section*{Appendix}
\addcontentsline{toc}{section}{Appendix}
\setcounter{section}{0}
\renewcommand{\thesection}{\arabic{section}}
\renewcommand{\theHsection}{appendix.\arabic{section}}
\raggedbottom
\section{Cost and Penalty Parameters}

The simulator uses the following fixed cost and value parameters.  All values are expressed in local currency units (LCU). The corresponding dollar values are also reported using Purchasing Power Parity (PPP) \cite{wbppp}. The simulator itself uses the LCU values during the experimentation, we convert the values post-experimentation into dollars.

\label{app:params}
\begin{table}[htbp]
\centering
\caption{Cost and penalty parameters used in Optimization Agent (OA).}
\label{tab:params}
\small
\renewcommand{\arraystretch}{1.15}
\begin{tabular}{@{}p{4.4cm}cp{1.5cm}p{1.5cm}p{6.0cm}@{}}
\toprule
Parameter & Notation & LCU & \$ & Source/Basis \\
\midrule

Grain price (per kg)
& $p_g$
& 20.90
& 1.04
& Support price \cite{cacp23} \\

UreaN (per kg N)
& $p_1^F$
& 13.04
& 0.65
& Retail price \cite{dof} \\

DAP (per kg)
& $p_2^F$
& 27.00
& 1.34
& Retail price \cite{dof} \\

Mulch (per kg)
& $p_M$
& 2.00
& 0.10
& Residue and fodder prices \cite{stover} \\

Water (per mm)
& $p_w$
& 2.50
& 0.12
& Irrigation charges \cite{itp,compmaize} \\

Labour (per operation)
& $p_L$
& 350.00
& 17.42
& Rural wage \cite{ls456} \\

Waterlogging (per ha-season)
& $S_{\mathrm{wl}}$
& 5{,}000
& 248.88
& Drainage cost \cite{drain,waterlogging_damage,singh13} \\

N oversupply (per ha)
& $S_{\mathrm{no}}$
& 2{,}000
& 99.55
& Excess-N penalty \cite{ina} \\

Deep drainage (per mm)
& $S_{\mathrm{dd}}$
& 15.00
& 0.75
& Water-loss and drainage impacts \cite{drain,motarjemi2023drainage} \\

N leaching (per kg N)
& $S_{\mathrm{nl}}$
& 25.00
& 1.24
& Nutrient-loss and environmental impacts
\cite{ina,zhang2023fertigation,zhu2005water,motarjemi2023drainage} \\

SOC loss (per kg C)
& $S_{\mathrm{cl}}$
& 150.00
& 7.47
& Social cost of carbon \cite{lee2024monetizing,scc} \\

Crop failure (per ha)
& $S_{\mathrm{cf}}$
& 50{,}000
& 2{,}488.80
& Cost of production \cite{cacp23} \\

\bottomrule
\end{tabular}
\end{table}

\subsection{Price Parameters}

The price parameters are based on prevailing market, administered, or farmer-paid prices reported in the literature and official sources. Where applicable, these prices reflect the prevailing subsidy and pricing policies in the country of study during the corresponding period. Thus, the parameters represent the effective prices relevant to the production decision.

\paragraph{Grain ($p_g$).}
The maize support price for 2023-24 was 2{,}090 LCU per quintal (100 kg) \cite{cacp23}, giving a value of 20.90 LCU per kg. 

\paragraph{UreaN ($p_1^F$).}
The reported farmer-paid price of a 45 kg Urea bag is 266.5 LCU and the bag contains 46\% N \cite{dof}. We express the N fertilizer cost on an 100 \% elemental-N basis and use 13.04 LCU per kg of N in the simulator, reflecting the effective fertilizer price under the prevailing pricing and subsidy system.

\paragraph{DAP ($p_2^F$).}
A 50 kg bag of DAP is reported at 1{,}350 LCU \cite{dof}, giving an effective price of 27.00 LCU per kg. This reflects the prevailing farmer-paid price under the applicable fertilizer pricing system.

\paragraph{Labour ($p_L$).}
Labour is charged as one person-day per operation. The reported rural wage was 316 LCU in 2023-24 and 349 LCU in 2024-25 \cite{ls456}. We use 350 LCU per operation.

\paragraph{Water ($p_w$).}
Reported irrigation charges for maize are approximately 254 LCU per ha \cite{compmaize}. Assuming about 100 mm of irrigation per season gives approximately 2.54 LCU per mm. We use 2.50 LCU per mm as the effective farmer-paid water cost.

\paragraph{Mulch ($p_M$).}
Stover has been reported at approximately 3.8 LCU per kg \cite{stover}. We use 2.00 LCU per kg as the effective value of crop residue retained in the field.

\subsection{Environmental Parameters}

The environmental parameters convert simulated resource losses into monetary values.

\paragraph{Waterlogging management ($S_{\mathrm{wl}}$).}
Waterlogging can reduce crop productivity and cause economic losses through yield losses and soil degradation. Reported estimates of waterlogging-related production losses range from 4{,}400 to 9{,}400 LCU per ha for individual crops, with a reported annual loss of 13{,}800 LCU per ha \cite{waterlogging_damage,singh13,drain}. We use 5{,}000 LCU per ha-season as a conservative partial-cost parameter for waterlogging.

\paragraph{Deep drainage ($S_{\mathrm{dd}}$).}
Deep drainage represents water that moves below the crop root zone and is therefore no longer available for crop uptake. Excessive deep drainage can also transport nutrients below the root zone, increasing nutrient losses and reducing nutrient-use efficiency \cite{motarjemi2023drainage,zhu2005water}. Such losses can reduce crop productivity and may require additional inputs to maintain crop growth. We use 15.00 LCU per mm of deep drainage as a model parameter representing the combined economic effect of excess water and associated nutrient losses.

\paragraph{N leaching ($S_{\mathrm{nl}}$).}
Nitrogen loss through leaching reduces nitrogen-use efficiency and can affect crop productivity while contributing to environmental pollution \cite{ina,zhang2023fertigation, zhu2005water,motarjemi2023drainage}. We use 25.00 LCU per kg of leached N as a conservative model parameter representing the economic penalty associated with nitrogen loss.

\paragraph{N oversupply ($S_{\mathrm{no}}$).}
The N oversupply parameter assigns a penalty to nitrogen applied beyond the crop's estimated nitrogen requirement. A value of 2{,}000 LCU is used as the penalty parameter, reflecting the economic and environmental costs associated with excess nitrogen application \cite{ina}.

\paragraph{SOC loss ($S_{\mathrm{cl}}$).}
SOC loss represents a reduction in carbon stored in the soil. Declining SOC can reduce soil fertility, water-holding capacity, and nutrient availability, thereby affecting crop productivity in subsequent seasons. It also represents a loss of carbon from the soil that may contribute to atmospheric CO$_2$ emissions. We therefore assign an economic value using the social cost of carbon. The global median social cost of carbon is estimated at US\$417 per tonne of CO$_2$ \cite{scc, lee2024monetizing}. Converting this value from CO$_2$ to elemental carbon and into LCU gives approximately 146 LCU per kg of C. We use 150 LCU per kg of C as the SOC-loss parameter.

\paragraph{Crop failure.}
The projected cost of maize production in 2023--24 was 1{,}394 LCU per quintal, with a projected yield of 36.96 q/ha \cite{cacp23}. This gives approximately 51{,}522 LCU per ha. We use 50{,}000 LCU per ha as the crop-failure parameter.

\section{Case Study: Full Weekly Advisories}
\label{app:case}
This section provides the natural language advisories provided by Agri-SCHOLAR during the 2023 season in the irrigated region.
\begin{multicols}{2}
\begin{advisory}
# PRE-SOWING (land preparation)
sow:     04 Jul 2023 (monsoon set in; soil moist enough
         for germination)
basal:   200 kg/ha DAP (4 bags of 50 kg) in the
         seed furrow at sowing
mulch:   26 kg/ha residue/straw between rows to
         save soil moisture
tillage: light/reduced tillage (20% surface disturbed)

# WEEK 0 | 04 Jul | 0 DAS
stage:   sowing / germination
weather: rain since 1 Jun = 88 mm; longest dry spell = 10 d
action:  apply 128.3 kg/ha of urea (2.85 bags of 45 kg)
         as side-dressing near plant rows on moist
         soil; cover with soil
why:     starter urea with basal DAP supports early root
         and leaf growth; model check: no water or N
         stress expected under this plan

# WEEK 1 | 11 Jul | 7 DAS
stage:   emergence / establishment
weather: rain since 1 Jun = 153 mm (65 mm past week);
         longest dry spell = 10 d
action:  no fertiliser or irrigation this week; continue
         routine scouting (weeds, fall armyworm, leaf blight)
why:     65 mm rain last week kept soil moist; week-0 urea
         still available, extra urea now would mostly
         leach; model check: no water or N stress
         expected under this plan

# WEEK 2 | 18 Jul | 14 DAS
stage:   seedling to early vegetative (V2-V6)
weather: rain since 1 Jun = 189 mm (36 mm past week);
         longest dry spell = 10 d
action:  no fertiliser or irrigation this week; continue
         routine scouting (weeds, fall armyworm, leaf blight)
why:     36 mm rain last week kept soil moist; week-0 urea
         still available, extra urea now would mostly
         leach; model check: no water or N stress
         expected under this plan

# WEEK 3 | 25 Jul | 21 DAS
stage:   seedling to early vegetative (V2-V6)
weather: rain since 1 Jun = 290 mm (101 mm past week);
         longest dry spell = 10 d
action:  no fertiliser or irrigation this week; continue
         routine scouting (weeds, fall armyworm, leaf blight)
why:     101 mm rain last week kept soil moist; week-0 urea
         still available, extra urea now would mostly
         leach; model check: no water or N stress
         expected under this plan

# WEEK 4 | 01 Aug | 28 DAS
stage:   seedling to early vegetative (V2-V6)
weather: rain since 1 Jun = 332 mm (42 mm past week);
         longest dry spell = 10 d
action:  apply 210.9 kg/ha of urea (4.69 bags of 45 kg)
         as side-dressing near plant rows on moist
         soil; cover with soil
why:     crop is entering its fastest growth phase and
         takes up most of its N now; top-dressing avoids
         leaf yellowing; model check: no water or N
         stress expected under this plan

# WEEK 5 | 08 Aug | 35 DAS
stage:   knee-high, rapid vegetative growth (V8-V12)
weather: rain since 1 Jun = 344 mm (12 mm past week);
         longest dry spell = 10 d
action:  no fertiliser or irrigation this week; continue
         routine scouting (weeds, fall armyworm, leaf blight)
why:     only 12 mm rain last week, but soil still holds
         enough moisture; week-4 urea still available,
         extra urea now would mostly leach; model check:
         no water or N stress expected under this plan

# WEEK 6 | 15 Aug | 42 DAS
stage:   knee-high, rapid vegetative growth (V8-V12)
weather: rain since 1 Jun = 351 mm (7 mm past week);
         longest dry spell = 11 d
action:  one irrigation of 63 mm (631,000 L/ha;
         roughly one furrow irrigation)
why:     soil moisture is running low for the current
         crop stage (dry spells of up to 11 d this season);
         model check: minor stress remains (water 5.4 d,
         N 0.0 d)

# WEEK 7 | 22 Aug | 49 DAS
stage:   pre-tasselling (V14-VT)
weather: rain since 1 Jun = 357 mm (6 mm past week);
         longest dry spell = 18 d
action:  apply 52.2 kg/ha of urea (1.16 bags of 45 kg)
         as side-dressing near plant rows on moist
         soil; cover with soil
why:     urea top-up keeps the crop green through the
         reproductive phase; model check: minor stress
         remains (water 4.7 d, N 0.0 d)

# WEEK 8 | 29 Aug | 56 DAS
stage:   tasselling and silking (most water-sensitive)
weather: rain since 1 Jun = 362 mm (5 mm past week);
         longest dry spell = 23 d
action:  no fertiliser or irrigation this week; continue
         routine scouting (weeds, fall armyworm, leaf blight)
why:     only 5 mm rain last week, but soil still holds
         enough moisture; week-7 urea still available,
         extra urea now would mostly leach; model check:
         minor stress remains (water 5.6 d, N 0.0 d)

# WEEK 9 | 05 Sep | 63 DAS
stage:   tasselling and silking (most water-sensitive)
weather: rain since 1 Jun = 382 mm (20 mm past week);
         longest dry spell = 23 d
action:  no fertiliser or irrigation this week; continue
         routine scouting (weeds, fall armyworm, leaf blight)
why:     20 mm rain last week kept soil moist; week-7 urea
         still available, extra urea now would mostly
         leach; model check: no water or N stress
         expected under this plan

# WEEK 10 | 12 Sep | 70 DAS
stage:   tasselling and silking (most water-sensitive)
weather: rain since 1 Jun = 394 mm (12 mm past week);
         longest dry spell = 23 d
action:  no fertiliser or irrigation this week; continue
         routine scouting (weeds, fall armyworm, leaf blight)
why:     only 12 mm rain last week, but soil still holds
         enough moisture; week-7 urea still available,
         extra urea now would mostly leach; model check:
         no water or N stress expected under this plan

# WEEK 11 | 19 Sep | 77 DAS
stage:   grain filling (milk to dough)
weather: rain since 1 Jun = 403 mm (9 mm past week);
         longest dry spell = 23 d
action:  no fertiliser or irrigation this week; continue
         routine scouting (weeds, fall armyworm, leaf blight)
why:     only 9 mm rain last week, but soil still holds
         enough moisture; week-7 urea still available,
         extra urea now would mostly leach; model check:
         no water or N stress expected under this plan

# WEEK 12 | 26 Sep | 84 DAS
stage:   grain filling (milk to dough)
weather: rain since 1 Jun = 422 mm (19 mm past week);
         longest dry spell = 23 d
action:  no fertiliser or irrigation this week; continue
         routine scouting (weeds, fall armyworm, leaf blight)
why:     19 mm rain last week, soil still holds enough
         moisture; week-7 urea still available, extra
         urea now would mostly leach; model check: no
         water or N stress expected under this plan

# WEEK 13 | 03 Oct | 91 DAS
stage:   grain filling (milk to dough)
weather: rain since 1 Jun = 444 mm (22 mm past week);
         longest dry spell = 23 d
action:  no fertiliser or irrigation this week; continue
         routine scouting (weeds, fall armyworm, leaf blight)
why:     22 mm rain last week kept soil moist; week-7 urea
         still available, extra urea now would mostly
         leach; model check: minor stress remains
         (water 2.1 d, N 0.0 d)

# WEEK 14 | 10 Oct | 98 DAS
stage:   dough to dent, grain hardening
weather: rain since 1 Jun = 446 mm (2 mm past week);
         longest dry spell = 23 d
action:  no fertiliser or irrigation this week; continue
         routine scouting (weeds, fall armyworm, leaf blight)
why:     crop is maturing, so fertiliser or water will no
         longer raise yield; model check: minor stress
         remains (water 1.6 d, N 0.0 d)

# WEEK 15 | 17 Oct | 105 DAS
stage:   dough to dent, grain hardening
weather: rain since 1 Jun = 451 mm (5 mm past week);
         longest dry spell = 23 d
action:  no fertiliser or irrigation this week; continue
         routine scouting (weeds, fall armyworm, leaf blight)
why:     crop is maturing, so fertiliser or water will no
         longer raise yield; model check: minor stress
         remains (water 2.8 d, N 0.0 d)

# WEEK 16 | 24 Oct | 112 DAS
stage:   physiological maturity / harvest preparation
weather: rain since 1 Jun = 451 mm (0 mm past week);
         longest dry spell = 23 d
action:  no fertiliser or irrigation this week; continue
         routine scouting (weeds, fall armyworm, leaf blight)
why:     crop is maturing, so fertiliser or water will no
         longer raise yield; model check: minor stress
         remains (water 2.9 d, N 0.0 d)

# WEEK 17 | 31 Oct | 119 DAS
stage:   physiological maturity / harvest preparation
weather: rain since 1 Jun = 465 mm (14 mm past week);
         longest dry spell = 27 d
action:  no fertiliser or irrigation this week; continue
         routine scouting (weeds, fall armyworm, leaf blight)
why:     crop is maturing, so fertiliser or water will no
         longer raise yield; model check: minor stress
         remains (water 2.9 d, N 0.0 d)

# SEASON SUMMARY
inputs:  urea 391.4 kg/ha (8.7 bags of 45 kg) in 3 splits
         (wk 0, 4, 7); irrigation 63 mm (1 event, wk 6);
         rest from rain
yield:   11,028 kg/ha (110.3 q/ha), simulated
cost:    $500.65/ha (simulated input cost)
leached: 12.9 kg N/ha nitrate
\end{advisory}
\end{multicols}

\providecommand{\bad}[1]{\textcolor{red}{[#1]}}
\providecommand{\badb}[1]{\textcolor{red}{\textbf{[#1]}}}   

\section{Ablation of the Advisory Bounding Agent}
\label{app:ablation}

This section gives the full results of the ablation in Section~\ref{sec:aba_section}. We compare the Agri-SCHOLAR pipeline (\textbf{With ABA}) against two variants in which the ABA is removed and an Optuna-based OA searches on its own:
\begin{itemize}
    \item \textbf{Loose Bounds}: the optimizer is limited only to fixed, numerically feasible ranges for each action, without the agronomic bounds the ABA derives from the farm context and retrieved literature.
    \item \textbf{No Bounds}: even these ranges are removed, so fertilizer and irrigation amounts have no upper limit.
\end{itemize}
The question is whether a purely profit-driven optimizer, given the same simulator and reward, can find plans a farmer could actually follow. All results are for the irrigated region over 2015--2025, and every plan is evaluated in APSIM under the true weather of the season.

In Tables~\ref{tab:ablation_yield} to~\ref{tab:ablation_leaching}, values for Loose Bounds and No Bounds are shown in \textcolor{red}{red square brackets}. The brackets mark outcomes of agronomically infeasible advisories (Appendix~\ref{app:ablation-obs}). They are what the simulator predicts, these are reported only to show how far an unconstrained optimizer can push the simulated metrics but these are not achievable.

\subsection{Results}
\label{app:ablation-results}

\paragraph{Loose Bounds.} On simulated yield and profit alone, Loose Bounds is competitive with Agri-SCHOLAR. It has the highest yield in 4 of 11 seasons and the highest net profit in 7 seasons, with its largest gains in 2016 and 2025 (over 3{,}000~kg\,ha$^{-1}$ above Agri-SCHOLAR). Environmental totals are similar: Agri-SCHOLAR has the lowest drainage in 6 seasons and the lowest leaching in 5, against 3 and 6 for Loose Bounds. 

The totals are weather-dependent, when heavy rain follows a large early dose, losses are pronounced: in 2017 Loose Bounds leached 30.73~kg\,ha$^{-1}$ of nitrate against 5.80 for Agri-SCHOLAR (over five times as much), and in 2023 it leached 22.52 against 12.90. Across all seasons, its sustainability penalty is 20.1\% higher than that of Agri-SCHOLAR (Table~\ref{tab:ablation}).

\paragraph{No Bounds.} Removing all bounds is far worse. Net profit is negative in four seasons (2019, 2022, 2024, 2025) and ranges from $-$\$2{,}887 to \$9{,}208~ha$^{-1}$. No Bounds reaches the highest yield in 2015 and 2024, but severe penalty costs leave its profit well below that of the bounded variants. Nitrate leaching is the highest of the three variants in every season, reaching 694.18~kg\,ha$^{-1}$ in 2025 and 3{,}270.66~kg\,ha$^{-1}$ in 2019. Deep drainage exceeds 800~mm in 2019 and 1{,}000~mm in 2020. It comes close to the bounded variants only in seasons such as 2023, where the realized weather happened to suit its large early applications.

\begin{table}[ht]
\begin{minipage}[t]{0.48\linewidth}
\centering
\caption{Yield (kg/ha) per season: Agri-SCHOLAR vs.\ without ABA (irrigated region, 2015--2025). Highest value per season in bold. \textcolor{red}{Red values in [\,]} come from agronomically infeasible advisories.}
\label{tab:ablation_yield}
\small
\setlength{\tabcolsep}{4pt}
\begin{tabular}{lrrr}
\toprule
& & \multicolumn{2}{c}{w/o ABA} \\
\cmidrule(l){3-4}
Year & With ABA & Loose Bounds & No Bounds \\
\midrule
2015 & 9,649 & \bad{10,070} & \badb{10,170} \\
2016 & 5,819 & \badb{9,200} & \bad{9,022} \\
2017 & \textbf{8,798} & \bad{5,419} & \bad{4,979} \\
2018 & 6,945 & \badb{7,162} & \bad{4,208} \\
2019 & \textbf{6,927} & \badb{6,927} & \badb{6,927} \\
2020 & \textbf{6,401} & \bad{4,679} & \bad{4,356} \\
2021 & \textbf{9,031} & \bad{7,667} & \bad{7,968} \\
2022 & 7,854 & \badb{8,127} & \bad{7,181} \\
2023 & \textbf{11,028} & \bad{10,544} & \bad{10,405} \\
2024 & 4,524 & \bad{5,805} & \badb{6,467} \\
2025 & 5,095 & \badb{8,163} & \bad{6,226} \\
\bottomrule
\end{tabular}
\end{minipage}\hfill
\begin{minipage}[t]{0.48\linewidth}
\centering
\caption{Net profit (\$/ha) per season: Agri-SCHOLAR vs.\ without ABA (irrigated region, 2015--2025). Highest value per season in bold. \textcolor{red}{Red values in [\,]} come from agronomically infeasible advisories.}
\label{tab:ablation_reward}
\small
\setlength{\tabcolsep}{4pt}
\begin{tabular}{lrrr}
\toprule
& & \multicolumn{2}{c}{w/o ABA} \\
\cmidrule(l){3-4}
Year & With ABA & Loose Bounds & No Bounds \\
\midrule
2015 & 9,470.27 & \badb{9,966.65} & \bad{3,733.94} \\
2016 & 5,295.06 & \badb{8,883.95} & \bad{2,188.71} \\
2017 & \textbf{8,705.98} & \bad{4,831.18} & \bad{2,767.43} \\
2018 & 6,509.34 & \badb{6,788.55} & \bad{2,320.83} \\
2019 & 6,571.44 & \badb{6,581.54} & \bad{$-$2,887.00} \\
2020 & \textbf{5,792.05} & \bad{4,231.49} & \bad{1,045.02} \\
2021 & \textbf{8,633.07} & \bad{7,311.67} & \bad{1,287.78} \\
2022 & 7,338.90 & \badb{7,724.62} & \bad{$-$103.31} \\
2023 & \textbf{10,745.40} & \bad{9,857.63} & \bad{9,208.20} \\
2024 & 3,753.07 & \badb{5,324.62} & \bad{$-$750.11} \\
2025 & 4,461.40 & \badb{7,696.51} & \bad{$-$301.95} \\
\bottomrule
\end{tabular}
\end{minipage}
\end{table}

\begin{table}[ht]
\begin{minipage}[t]{0.48\linewidth}
\centering
\caption{Cumulative deep drainage (mm) per season: Agri-SCHOLAR vs.\ without ABA (irrigated region, 2015--2025). Lowest value per season in bold. \textcolor{red}{Red values in [\,]} come from agronomically infeasible advisories.}
\label{tab:ablation_drainage}
\small
\setlength{\tabcolsep}{4pt}
\begin{tabular}{lrrr}
\toprule
& & \multicolumn{2}{c}{w/o ABA} \\
\cmidrule(l){3-4}
Year & With ABA & Loose Bounds & No Bounds \\
\midrule
2015 & 45.00 & \bad{42.63} & \badb{42.61} \\
2016 & 282.00 & \badb{228.95} & \bad{233.58} \\
2017 & \textbf{84.00} & \bad{297.67} & \bad{386.83} \\
2018 & \textbf{197.00} & \bad{201.08} & \bad{569.57} \\
2019 & \textbf{194.00} & \bad{202.90} & \bad{818.60} \\
2020 & 382.00 & \badb{304.65} & \bad{1,089.61} \\
2021 & 291.00 & \bad{269.60} & \badb{228.84} \\
2022 & \textbf{299.00} & \bad{303.53} & \bad{339.13} \\
2023 & \textbf{151.00} & \bad{263.82} & \bad{342.66} \\
2024 & 407.00 & \badb{335.67} & \bad{340.41} \\
2025 & \textbf{349.00} & \bad{363.58} & \bad{598.73} \\
\bottomrule
\end{tabular}
\end{minipage}\hfill
\begin{minipage}[t]{0.48\linewidth}
\centering
\caption{Cumulative NO\textsubscript{3} leaching (kg/ha) per season: Agri-SCHOLAR vs.\ without ABA (irrigated region, 2015--2025). Lowest value per season in bold. \textcolor{red}{Red values in [\,]} come from agronomically infeasible advisories.}
\label{tab:ablation_leaching}
\small
\setlength{\tabcolsep}{4pt}
\begin{tabular}{lrrr}
\toprule
& & \multicolumn{2}{c}{w/o ABA} \\
\cmidrule(l){3-4}
Year & With ABA & Loose Bounds & No Bounds \\
\midrule
2015 & 3.10 & \badb{2.95} & \bad{3.11} \\
2016 & 23.70 & \badb{18.76} & \bad{76.44} \\
2017 & \textbf{5.80} & \bad{30.73} & \bad{33.60} \\
2018 & 16.40 & \badb{15.79} & \bad{54.83} \\
2019 & \textbf{17.90} & \bad{26.34} & \bad{3,270.66} \\
2020 & 34.70 & \badb{20.94} & \bad{275.83} \\
2021 & \textbf{28.70} & \bad{34.59} & \bad{102.15} \\
2022 & 28.10 & \badb{25.10} & \bad{296.62} \\
2023 & \textbf{12.90} & \bad{22.52} & \bad{36.55} \\
2024 & 40.10 & \badb{28.90} & \bad{364.21} \\
2025 & \textbf{36.50} & \bad{36.68} & \bad{694.18} \\
\bottomrule
\end{tabular}
\end{minipage}
\end{table}

\subsection{Observations}
\label{app:ablation-obs}

\paragraph{The optimizer exploits blind spots in the simulator and reward.}
\label{app:ablation-gaming}
Without the ABA, the optimizer treats APSIM as a black box and maximizes the reward, which is reliable only if the simulator and reward capture everything that matters in the field. They do not. APSIM does not model  the smallest dose that can be spread evenly, or the effort of a field operation beyond the labour cost. The reward penalizes leaching, deep drainage, waterlogging, and nitrogen oversupply, but has no term for whether a dose suits the crop stage or can be applied at all. Because each application carries a fixed labour cost ($p_L = 350$~LCU), the optimizer is drawn to two kinds of schedules that look efficient under the reward but are unsound on the farm.

\begin{itemize}
    \item \textbf{Excessive nitrogen at planting.} In 2019, the optimizer front-loaded nitrogen into a single application, Loose Bounds applied 190~kg\,ha$^{-1}$ of UreaN at planting, and No Bounds applied 6{,}062.2~kg\,ha$^{-1}$. A maize seedling has almost no root system and takes up little nitrogen, so such a dose sits in the soil exposed to leaching, and urea near germinating seed risks ammonia injury that APSIM never penalizes.
    \item \textbf{Doses too small to apply.} The optimizer also adds token top-ups of 2--4~kg\,ha$^{-1}$, such as 2.8~kg\,ha$^{-1}$ of UreaN at week~8 of 2024. These buy a marginal simulated yield gain while staying below the leaching penalty, but 2.8~kg is about 6\% of a single 45~kg bag. It cannot be calibrated or spread evenly with standard equipment, and no farmer would spend a person-day on it.
\end{itemize}

\paragraph{Simulated gains are not evidence of better advice.}
Large single doses are also a bet that the following weeks will not bring heavy rain, since the optimizer plans on the constructed weather available at advisory time. The 2019 No Bounds plan shows the failure: under the true weather it leached 3{,}270.66~kg\,ha$^{-1}$ of nitrate yet yielded 6{,}927~kg\,ha$^{-1}$, identical to both bounded variants. The extra nitrogen bought no yield and produced a net loss of \$2{,}887~ha$^{-1}$. Loose Bounds' apparent advantage in several seasons is likewise an artefact of a model and reward that cannot judge agronomic quality. The scores reflect skill at exploiting the model, not the quality of the advice.

\paragraph{The ABA closes these gaps.}
By restricting the search to doses and timings that agronomic guidance supports for the current crop stage, the ABA yields plans a farmer can follow. For example, In the 2023 case study (Appendix~\ref{app:case}), Agri-SCHOLAR splits 180~kg\,ha$^{-1}$ of UreaN into a starter dose at sowing (59~kg\,ha$^{-1}$), a main top-dress at rapid vegetative growth (97~kg\,ha$^{-1}$), and a top-up before tasselling (24~kg\,ha$^{-1}$). Each dose matches crop demand at that stage and can be applied in whole or half bags. The cost is a small loss in simulated profit in some seasons, in exchange for avoiding the leaching and profit losses seen when the bounds are removed.

\end{document}